\documentclass[10pt]{article} % For LaTeX2e
\usepackage[accepted]{tmlr}
\usepackage[breaklinks, colorlinks=true, citecolor=Blue, linkcolor=BrickRed, urlcolor=MidnightBlue]{hyperref}

\usepackage{textcomp}
\usepackage{stfloats}
\usepackage{url}
\usepackage{verbatim}
\usepackage{graphicx}
\usepackage[table]{xcolor}

\usepackage{amssymb}
\usepackage{booktabs}

\usepackage[dvipsnames, table]{xcolor}
\usepackage{subcaption}

\usepackage{pgfplots}
\usepackage{pgfplotstable}
\pgfplotsset{compat=newest}
\usetikzlibrary{arrows.meta}
\usetikzlibrary{shapes}
\usetikzlibrary{positioning}

\usepackage{comment}
\usepackage[tbtags]{mathtools}
\usepackage{physics}
\usepackage{nicefrac}

\usepackage{algorithm}
\usepackage{algpseudocodex}

\usepackage{makecell}

\usepackage{dirtytalk}
\usepackage[capitalize]{cleveref}
\usepackage[normalem]{ulem}
\usepackage{multirow}

\usepackage{enumitem}

\title{Maximising the Utility of Validation Sets for Imbalanced Noisy-label Meta-learning}
\author{\name Dung Anh Hoang \email hoang.dung@monash.edu \\
      \addr Department of Data Science and AI \\
      Monash University 
      \AND
      \name Cuong Nguyen \email c.nguyen@surrey.ac.uk \\
      \addr Centre for Vision, Speech and Signal Processing \\
    University of Surrey
      \AND
      \name Vasileios Belagiannis \email vasileios.belagiannis@fau.de\\
      \addr Friedrich-Alexander-Universität Erlangen-Nürnberg
      \AND
      \name Thanh-Toan Do \email toan.do@monash.edu\\
      \addr Department of Data Science and AI\\
      Monash University
      \AND
      \name G. Carneiro \email g.carneiro@surrey.ac.uk\\
      \addr Centre for Vision, Speech and Signal Processing \\
University of Surrey}

\def\month{03}  % Insert correct month for camera-ready version
\def\year{2025} % Insert correct year for camera-ready version
\def\openreview{\url{https://openreview.net/forum?id=SBM9yeNZz5}} % Insert correct link to OpenReview for camera-ready version

\begin{document}

\maketitle

\begin{abstract}
    \noindent
    Meta-learning is an effective method to handle imbalanced and noisy-label learning, but it generally depends on a clean validation set. Unfortunately, this validation set has poor scalability when the number of classes increases, as traditionally these samples need to be randomly selected, manually labelled and balanced-distributed. This problem therefore has motivated the development of meta-learning methods to automatically select validation samples that are likely to have clean labels and balanced class distribution. Unfortunately, a common missing point of existing meta-learning methods for noisy label learning is the lack of consideration for data informativeness when constructing the validation set. The construction of an informative  validation set requires hard samples, i.e., samples that the model has low confident prediction, but these samples are more likely to be noisy, which can degrade the meta reweighting process. Therefore, the balance between sample informativeness and cleanness is an important criteria  for validation set optimization.
    %that is clean, balanced, and informative for the meta-learning algorithm.
    % The random selection and manual labelling and balancing of this validation set is not only sub-optimal for meta-learning, but it also scales poorly with the number of classes.
    % Hence, recent meta-learning papers have proposed heuristics to automatically build and label this validation set, but these heuristics focus only on label cleanliness and balanced class distribution, without taking into account sample informativeness, which can be sub-optimal for meta-learning.
    In this paper, we propose new criteria  to characterise the utility of such meta-learning validation sets, based on: 
    1) sample informativeness; 2) balanced class distribution; and 3) label cleanliness.
    We also introduce a new imbalanced noisy-label meta-learning (INOLML) algorithm that automatically builds a validation set by maximising such utility criteria.
    The proposed method shows state-of-the-art (SOTA) results compared to previous meta-learning and noisy-label learning approaches on several noisy-label learning benchmarks.
    % \keywords{Meta Learning, Noisy-label Learning, Imbalanced Learning} % Active Learning
\end{abstract}

\section{Introduction\label{sec:introduction}}
    The development of new deep learning methods will gradually depend more on poorly curated datasets~\citep{clothing1M,Webvision}. Such datasets tend to have class-imbalanced distributions and to contain large amounts of noisy-label samples.
    The problems of class-imbalanced learning and noisy-label learning have, however, been addressed separately. 
    For instance, while noisy-label methods are based on robust loss functions~\citep{Robust_loss,wang2019symmetric}, label cleaning~\citep{jaehwan2019photometric,yuan2018iterative}, meta-learning~\citep{L2W,han2018pumpout}, ensemble learning~\citep{miao2015rboost}, 
    and semi-supervised learning~\citep{li2020dividemix},
    imbalanced learning approaches  often rely on meta-learning~\citep{L2W,han2018pumpout,FSR}, transfer learning~\citep{transfer_imb_2,transfer_imb}, classifier design~\citep{class_imb_2,class_imb}, and re-sampling~\citep{resample_2}. Notice from the methods above that meta-learning methods have the potential to address both noisy-label and imbalanced learning problems~\citep{L2W, Distill_noise, FSR, Famus, MetaWeightNetLA}.

    %Meta-learning is often formulated as an iterative bi-level optimization, where the lower level uses a training set to train a classification model with the meta-parameters estimated from the upper-level optimization (the meta parameters are usually represented by training sample weights or pseudo-label parameters). Then, the upper level optimises the meta parameters with the model obtained from the lower level using a validation set. To be effective against imbalanced noisy-label problems, the validation set must be clean and balanced, which is usually achieved through random selections and manually labelling.
    Meta-learning is often formulated as a bi-level optimization, where the upper-level optimization estimates the meta-parameters of interest using a validation set, while the lower-level optimization trains a model using a training set. In meta-learning, the validation set is commonly built by randomly selecting and manually labelling a class-balanced set of samples. 
    However, the process of building these validation sets scales poorly when the number of classes escalates. Such issue, therefore, motivates the design of methods that automatically build validation sets containing pseudo-clean and class-balanced samples~\citep{FSR,Famus}. However, the prediction accuracy results of such methods are lower than the accuracy from conventional approaches that rely on manually-curated validation sets~\citep{Distill_noise}. 
    We argue that this poor performance might be attributed to the low informativeness of the validation sets based on pseudo-clean samples~\citep{FSR,Famus},  comparing to the potentially more informative randomly selected samples. The reason is that these methods tend to prioritize only high-confidence samples, as they are more likely to be clean, but such samples are often less informative to the model.
    As a result, defining \emph{sample informativeness} within a meta reweighting framework emerges as a crucial challenge. At the same time, intuitively, \emph{informative samples} are more likely to be \emph{hard sample} (samples with low confident prediction, high gradient magnitude from the model), thus also more likely to be noisy, which can contaminate the validation set and degrade the performance of the meta learning framework. This issue has highlighted the need for a framework that can automatically construct and optimize the balance between informativeness and the integrity of the validation set.
    
    %accuracy can be attributed to their proposed heuristics~\citep{FSR,Famus}, which can characterise well balanced distribution and label cleanliness, but not sample informativeness. % for the meta-learning algorithm.
 %\vasilis{This paragraph clearly describes the setup to highlight the problem at the end, but it's too long. Would it be possible to shorten it by 1-2 sentences?}
    %However, this way of building the validation set has many limitations, such as: given that classification accuracy depends on a large and clean validation set, such process scales poorly with the number of classes; and there are no clear criteria on how to select informative samples to be included in this validation set.\vasilis{This paragraph clearly describes the setup to highlight the problem at the end, but it's too long. Would it be possible to shorten it by 1-2 sentences?}
    
     Motivated by the bi-level optimization of the meta-learning algorithm, we propose novel criteria to characterise the utility of validation sets used in meta-learning. In particular, the proposed criteria are based on: 1) sample informativeness; 2) class-balanced distribution; and 3) label cleanliness.
     We also introduce a new imbalanced noisy-label meta-learning (INOLML) method that automatically builds a validation set by maximising its utility according to our proposed criteria. The proposed method, depicted in Figure~\ref{fig:motivation}, consists of 3 iterative steps: 1) detecting
     pseudo-clean samples from the noisy training set and labelling them; 2) forming the validation set from the robustly-labelled pseudo-clean set in step (1), using the proposed utility criteria; and 3) meta-learning using the validation set from step (2). The main contributions of our paper can be summarised as follows:
     \begin{itemize}%[noitemsep,topsep=5pt]
        \item A new criteria set to form the meta-learning validation set based on sample informativeness, class-balanced distribution, and label cleanliness.
         \item An innovative meta-learning algorithm (Figure~\ref{fig:motivation}), which automatically builds a validation set by maximising its utility according to our criteria, comprising the following steps:
         1) detection and robust labelling of pseudo-clean samples from the noisy training set; 2)
         formation of the validation set using the proposed utility criteria; and 3) meta-learning using the validation set from step (2). 
         \item A state-of-the-art meta-learning method to form validation sets for meta-learning that outperforms prior approaches that rely on random selecting or manually labelling. %based on manual or automatic methods to build validation sets.
         %based on an iterative update (using the sample selection criteria above) of a small validation set (with no more than 10 samples per class), which is empirically shown to be more effective than the common non-iterative process based on randomly selecting and manually labelling a small validation set of the same size (also with no more than 10 samples per class)~\citep{Distill_noise}.
         %\vasilis{Our contribution is the validation set selection right? Here, it reads as we also propose a meta-learning algorithm. I suggest to have bullet points for the contributions on the validation set selection and the SOTA results.}
     \end{itemize}
     %With the technical contributions above, 
     %our method improves over previous meta-learning approaches and set new state-of-the-art results on 
     %Inspired by the the work of Ren et. al\citep{L2LWS}, our methods introduce an objective formula to select the best clean validation set based on the similarity of s embedding feature and gradients between the clean validation set and the training samples.
     In addition, the proposed meta-learning method INOLML shows competitive results with respect to previous meta-learning and noisy-label learning approaches on   noisy-label learning benchmarks. 
%     In balanced noisy-label benchmarks, our method is competitive or better than the state-of-the-art.
     % \vspace{-0.25em}

\section{Related Work}
%\vspace{-0.5em}
\label{sec:related_work}
    %We review methods that can deal with imbalanced noisy-label learning, focusing on meta-learning approaches.

    %\vspace{-1em}
    \subsection{Noisy-label learning}
    \label{sec:related_work_noisy_label_learning}
        \noindent
        Noisy-label learning methods rely on many strategies, namely: robust loss functions~\citep{loss_func_1,zhang2018generalized,ghosh2017robust}, 
        %ensemble learning \citep{miao2015rboost}, 
        %student-teacher model~\citep{tarvainen2017mean}, 
        label cleaning \citep{yuan2018iterative, jaehwan2019photometric}, 
        co-teaching~\citep{li2020dividemix, MentorNet, han2018co},
        %dimensionality reduction~\citep{ma2018dimensionality},
        iterative label correction~\citep{label_clean,arazo2019unsupervised,DMLP}, 
        semi-supervised learning~\citep{ortego2019towards, li2020dividemix, ortego2021multi}, 
        contrastive learning~\citep{twin_contrast},
        meta-learning ~\citep{L2W,Distill_noise,FSR,Famus,MetaWeightNetLA}, 
        and hybrid methods~\citep{SELF,jiang2020beyond}. Except for the meta-learning approaches~\citep{L2W,Distill_noise,FSR,Famus,MetaWeightNetLA}, the majority of these methods assume a class-balanced training dataset.
        % with a class balanced distribution.
        %, except for the meta-learning approaches~\citep{MentorNet,L2W,Distill_noise,FSR,Famus,MetaWeightNetLA} that not only address the noisy-label problem, but also the learning with an imbalanced training set.

        %Meta learning is a versatile solution for  many problems (few-shot learning, reinforcement learning, etc.) that optimises  meta-parameters in order to benefit the training process.
        For meta-learning approaches~\citep{L2W, Distill_noise, FSR, Famus, MetaWeightNetLA}, meta-parameters are introduced to automatically down-weight the losses of noisy samples and up-weight the losses of clean samples~\citep{L2LWS, CWS}. Such meta-parameters are then learnt by optimising the one-step-ahead model on a validation set. 
        
        % the meta-parameters consist of a weight for each training sample~\citep{FSR,Distill_noise}, and the meta-learning optimises the model of interest based on a weighted cross-entropy loss that automatically downweights noisy samples and upweights clean samples~\citep{L2LWS,CWS}.
        % %For example, L2LWS~\citep{L2LWS} and CWS~\citep{CWS} comprise a target deep neural network (DNN) and a meta-DNN that is pre-trained on a small clean validation dataset to re-weight the training samples to model the target DNN.
        % Automatic re-weighting~\citep{L2W} weights training samples based on the performance of one-step-ahead model on the validation set. 
        To deal with noisy labels effectively, most of existing meta-learning approaches require a clean validation set. Such a set is, however, expensive to acquire and does not scale well with the number of classes. Recent studies relax such assumption by selecting the validation set from the noisy training set~\citep{FSR,Famus}. Unfortunately, their sample selection process demonstrate lower performance compared to their counterparts using real validation data. %has not been well-studied. 
        This motivates us to develop an approach that can automatically build a clean and balanced validation set, but unlike them, our approach is motivated by the meta-learning optimization that also takes into consideration the informativeness of the validation samples.

    \subsection{Imbalanced learning}
    \label{sec:related_work_imbalanced_learning}
        %Imbalance learning is another challenging classification problem that is commonly present in real-world datasets,  where a small portion of majority classes have a massive amount of training samples, and minority classes only have a few training samples~\citep{longtail_survey}. 
        Imbalanced learning can result in a biased model with good accuracy for majority classes, but poor performance for the minority ones~\citep{longtail_survey}.
        Methods to address imbalanced learning are based on transfer learning~\citep{transfer_imb_2, transfer_imb}, classifier design~\citep{class_imb, class_imb_2}, cost-sensitive learning~\citep{cost_sensitive_2, cost_sensitive_3, cost_sensitive}, data augmentation~\citep{data_aug, data_aug_2}, logit adjustment~\citep{logit_adjust, logit_adjust_2} and representation learning~\citep{represent_2, represent}.
        %Meta-learning is considered a branch of re-sampling methodologies~\citep{resample_2}. 
        These existing methods often assume that training labels are clean, which does not hold for noisy-label datasets.
        %\vspace{-4.5em}
        %\vspace{-2em}
        
    \subsection{Noisy-label and imbalanced learning}
        Most of the methods mentioned in \cref{sec:related_work_noisy_label_learning,sec:related_work_imbalanced_learning} treat noisy-label and imbalance learning as two separate problems.
        %, except for FSR~\citep{FSR}, which is an approach based on meta-learning.
        Label noise in imbalanced datasets has also been considered by non meta-learning approaches~\citep{Imbalance_noisy, ROLT, Karthik2021LearningFL}, but they either have different setups or achieve sub-par results. 
        For meta-learning, the validation set is crucial to allow good model performance, where we empirically observe that different randomly selected clean validation sets can lead to substantially different performances. 
        For example, if the validation set only contains \say{easy samples} that lie far from classification boundaries, meta-learning tends to produce poor classification accuracy. 
        Unfortunately, previous approaches have not studied this issue.
        For instance, classic meta-learning approaches~\citep{L2W,MetaWeightNetLA} rely on random sample selection and manual labelling, potentially producing uninformative validation sets. 
        Furthermore, the fact that such manually-curated validation set is fixed for the whole training process may hinder model generalisation because the meta-learning optimization can quickly overfit to this validation set. 
        Recent meta-learning approaches try to automatically build clean validation sets
        with \say{easy samples}, obtained from 
        low-loss samples~\citep{Famus} or well-optimised samples~\citep{FSR}.
        %. For instance, FaMUS~\citep{Famus} selects training samples with low losses to form the validation set, while FSR~\citep{FSR} chooses samples that can be well optimised after a training iteration given that such samples are more likely to be clean.
        Instead of selecting random samples or \say{easy samples}, we study how to select informative samples for meta-learning.
%        However, these methods aim to find the "easy samples" instead of informative samples for training, which may not be ideal for meta-learning.
        % maximise the cleanness of the validation set without considering their informativeness. 
        % To address this issue, we aim to proposed meta-learning methods which build validation sets that also contain informative samples.
        %, which is the problem being studied in this paper. 
        %\dung{I explain more motivation here}

        % \vspace{-1em}

\section{Background}
\label{sec:background}
    \noindent
    This section presents an overview of the label noise problem and two meta-learning methods widely-used when dealing with noisy-label datasets. %After that is our contribution for their improvement.

    \subsection{Noisy-label learning}
    \label{sec:noisy_label_learning}
        In the conventional supervised learning, we are given a training set \(\mathcal{D}_{\mathrm{clean}} = \{(\mathbf{x}_{i}, \mathbf{t}_{i})\}_{i=1}^{|\mathcal{D}_{\mathrm{clean}}|}\), where $\mathbf{x}_{i} \in \mathcal{X} \subseteq \mathbb{R}^{D}$ represents a \(D\)-dimensional input data, and \(\mathbf{t}_{i} \in \mathcal{Y} = \{\mathbf{t}: \mathbf{t} \in \{0, 1\}^{C} \wedge \pmb{1}^{\top} \mathbf{t} = 1 \}\) is a \(C\)-dimensional one-hot clean label. The aim is to find a classification model \(f_{\theta}: \mathcal{X} \to \Delta_{C - 1}\), parameterised by $\theta \in \Theta$, that maps the input data to its corresponding label, where \(\Delta_{C - 1} = \{\mathbf{p}: \mathbf{p} \in [0, 1]^{C} \wedge \pmb{1}^{\top} \mathbf{p} = 1\}\) denotes the \((C-1)\)-dimensional probability simplex.

        In practice, instead of observing the correctly-labelled training set \(\mathcal{D}_{\mathrm{clean}}\), we are given a noisy training set \({\mathcal{D} = \{(\mathbf{x}_{i}, \mathbf{y}_{i})\}_{i=1}^{|\mathcal{D}|}}\), where the noisy label \(\mathbf{y}_{i} \in \mathcal{Y}\) might or might not be the same as the clean label \(\mathbf{t}_{i}\). The aim is to exploit the information in such noisily-annotated training set to still learn a good model \(f_{\theta}\) that can accurately predict the clean label \(\mathbf{t}\) for an instance \(\mathbf{x}\).

    \subsection{Learning to reweight}
    \label{sec:background_learning_to_reweight}
    \emph{Learning to reweight}~\citep{L2W} is a meta-learning approach proposed to address the label noise problem defined in \cref{sec:noisy_label_learning}. Given the nature of meta-learning, the original training set is split into two non-overlapping subsets: training (or support) set \(\mathcal{D}^{(t)}\) and validation (or query) set \(\mathcal{D}^{(v)}\). The main idea is to employ the validation subset \(\mathcal{D}^{(v)}\) to learn a meta-parameter \(\omega \in \Delta_{|\mathcal{D}^{(t)}| - 1}\) (in the form of a probability vector) that weights the cross-entropy loss \(\ell\) of each training sample in the training subset \(\mathcal{D}^{(t)}\). The process obtaining \(\omega\) is based on maximising some utility criteria, e.g., informativeness, class-balanced distribution and label cleanliness~\citep{L2W}. Formally, the objective function of this meta-learning approach is a bi-level optimization defined as:
    % \begin{equation}
    %     \begin{aligned}[b]
    %         & \omega^{*} =  \arg\min_{\omega,\lambda} \frac{1}{|\mathcal{D}^{(v)}|} \sum_{(\mathbf{x}_j, \mathbf{y}_j) \in \mathcal{D}^{(v)}} \ell^{(v)} \left( \mathbf{x}_j, \mathbf{y}_j; \theta^{*} \left( \omega \right) \right) \\ % + \chi \\
    %         & \text{s.t.: }
    %         \theta^{*}(\omega) = \arg\min_{\theta}             \frac{1}{\abs{\mathcal{D}^{(t)}}}
    %         \sum_{( \mathbf{x}_i, \mathbf{y}_i ) \in \mathcal{D}^{(t)}}
    %           \ell^{(t)} \left(\mathbf{x}_i, \mathbf{y}_i,\omega_{i}; \theta \right),
    %     \end{aligned}
    %     \label{eq:traditional_meta_learning_bilevel_optimization}
    % \end{equation}
    % where \(\ell^{(v)}(\mathbf{x}_j,\mathbf{y}_j;\theta^{*}(\omega)) = \ell_{\mathrm{CE}}(\mathbf{y}_j, f_{\theta^{*}(\omega)}(\mathbf{x}_j))\) is the cross-entropy (CE) loss over the validation set, while $\ell^{(t)} \left(\mathbf{x}_i, \mathbf{y}_{i},\omega_{i}; \theta \right)=\omega_{i} \ell_{\mathrm{CE}}(\mathbf{y}_j, f_{\theta}(\mathbf{x}_j))$ is the weight CE loss using the training samples and the meta weight $\omega_{i}$.
    \begin{equation}
        \begin{aligned}[b]
            & \min_{\omega} \mathbb{E}_{(\mathbf{x}_{j}, \mathbf{y}_{j}) \in \mathcal{D}^{(v)}} \left[ \ell \left( f_{\theta^{*}(\omega)}(\mathbf{x}_{j}), \mathbf{y}_{j} \right) \right] \quad \text{s.t.: } \theta^{*}(\omega) = \arg\min_{\theta} \mathbb{E}_{(\mathbf{x}_{i}, \mathbf{y}_{i}) \in \mathcal{D}^{(t)}} \left[ \omega_{i} \ell \left( f_{\theta}(\mathbf{x}_{i}), \mathbf{y}_{i} \right) \right],
        \end{aligned}
        \label{eq:meta_learning_learn2reweight}
    \end{equation}
    where the notation \(\mathbb{E}\) denotes the expectation and can be expressed as \(\mathbb{E}_{\mathbf{u} \in \mathcal{U}} \left[ g(\mathbf{u}) \right] = \nicefrac{1}{|\mathcal{U}|} \sum_{\mathbf{u}_{i} \in \mathcal{U}} g(\mathbf{u}_{i})\).

    Intuitively, the lower-level (or the constraint) in \eqref{eq:meta_learning_learn2reweight} learns a classifier with weighted cross-entropy loss on the training subset \(\mathcal{D}^{(t)}\), while the upper-level evaluate the performance of the classifier on the training validation subset \(\mathcal{D}^{(v)}\) and optimises that performance w.r.t. the meta-parameter \(\omega\).

    The bi-level optimization in \eqref{eq:meta_learning_learn2reweight} can be solved by iterating the following two steps. In the first step, the parameter of the meta-learning model, \(\theta^{*}(\omega)\), is estimated from the stochastic gradient descent (SGD) on the training subset \(\mathcal{D}^{(t)}\), with each step defined by:
    \begin{equation}
    % \scalebox{0.8}{$
        \begin{aligned}
            & \theta^{*}(\omega) = \theta_{0} - \eta_{\theta} \grad_{\theta} \mathbb{E}_{( \mathbf{x}_{i}, \mathbf{y}_{i} ) \in \mathcal{D}^{(t)}} \omega_i \, \ell \left[ \left( f_{\theta}(\mathbf{x}_{i}), \mathbf{y}_{i} \right) \right],
            \end{aligned}
            % $}
        \label{eq:traditional_update_theta}
    \end{equation}
    where \(\eta_{\theta}\) is the step size or learning rate.

    In the second step, the meta-parameter \(\omega\) in the upper-level is updated by applying one SGD step on the loss evaluated on the validation subset \(\mathcal{D}^{(v)}\) using the classifier's parameter \(\theta^{*}(\omega)\) obtained in \eqref{eq:traditional_update_theta}. The update is defined as:
    \begin{equation}
        \begin{aligned}
            & {\omega}^{*} = \omega_{0} - \eta_{\omega} \grad_{\omega} \mathbb{E}_{(\mathbf{x}_{j}, \mathbf{y}_{j}) \in \mathcal{D}^{(v)}} \left[ \ell \left( f_{\theta^{*}(\omega)} (\mathbf{x}_{j}), \mathbf{y}_{j} \right) \right],
        \end{aligned}
        \label{eq:traditional_update_w}
    \end{equation}
    where \(\eta_{\omega}\) is the step size to update \(\omega\).

    The meta-parameter \(\omega^{*}\) obtained in \cref{eq:traditional_update_w} is projected to the \((|\mathcal{D}^{(t)}| - 1)\)-dimensional probability simplex, \(\Delta_{|\mathcal{D}^{(t)}| - 1}\), before being used to train the classifier of interest with the weighted loss defined in the lower-level of \eqref{eq:meta_learning_learn2reweight}.
    % \begin{equation}
    % % \scalebox{0.85}{$
    %      \begin{aligned}[b]
    %      \ell \left(\mathbf{x}_i, \mathbf{y}_i; \theta \right) = & \omega^{*}_i \ell_{\mathrm{CE}} \left(\mathbf{y}_{i}, f_{\theta}(\mathbf{x}_i) \right)
    %     \end{aligned}
    %     % $}
    %     \label{eq:traditional_training_loss_model_parameters}
    % \end{equation}

    \subsection{Meta-relabelling}
    \label{sec:meta_relabelling}
        Based on the \emph{learning to reweight} approach presented in \cref{sec:background_learning_to_reweight}, the \emph{Distill} method~\citep{Distill_noise} 
        proposes a slight modification of the objective function in \eqref{eq:meta_learning_learn2reweight} as follows:
        \begin{equation}
            \begin{aligned}[b]
                & \min_{\omega,\lambda} \mathbb{E}_{(\mathbf{x}_j, \mathbf{y}_j) \in \mathcal{D}^{(v)}} \left[ \ell \left( f_{\theta^{*} \left( \omega,\lambda \right)} (\mathbf{x}_j), \mathbf{y}_j \right) \right] \quad \text{s.t.: } \theta^{*}(\omega, \lambda) = \operatorname*{argmin}_{\theta}             \mathbb{E}_{( \mathbf{x}_i, \mathbf{y}_i ) \in \mathcal{D}^{(t)}} \left[ \omega_{i} \ell \left( f_{\theta}(\mathbf{x}_i), \hat{\mathbf{y}}(\lambda_i) \right) \right],
            \end{aligned}
            \label{eq:distill_supervision_label_noise}
        \end{equation}
        in which \(\lambda = \{\lambda_{i}\}_{i = 1}^{|\mathcal{D}^{(t)}|}\) is considered as another meta-parameter representing the tradeoff between the original label and model prediction. For each training sample \(\mathbf{x}_{i}\) in the training subset \(\mathcal{D}^{(t)}\), \emph{Distill} uses two labels,  $\hat{\mathbf{y}}_i(\lambda_0)$ for sample reweighting and $\mathbf{y}_i^{*}(\lambda^*_i)$ for sample relabeling. The modified label $\hat{\mathbf{y}}_i$ given the (noisy) label \(\mathbf{y}_{i}\) and trade-off variable $\lambda_i$ is defined as:
        \begin{equation}
            \hat{\mathbf{y}}_i(\lambda_i) = \lambda_i \mathbf{y}_i + (1-\lambda_i)f_{\theta}(\mathbf{x}_i),
            \label{eq:relabelling}
        \end{equation}
        where \(\lambda_{i} = 0.9 \quad \forall i={1,2,\dots,|\mathcal{D}^{(t)}|}\). For label $\hat{\mathbf{y}}_i(\lambda_i)$, instead of optimizing the parameter $\lambda_0$, its value is fixed at 0.9 (close to 1) to ensure that label $\hat{\mathbf{y}}_i(\lambda_i)$ retains information from the original label.  Supervised learning with label $\hat{\mathbf{y}}_i(\lambda_i)$ and the optimized weight $\omega_{i}$ effectively corresponds to upweighting clean samples and downweighting noisy samples. %\(\lambda_{i} \in [0, 1], i \in \{1, \ldots, |\mathcal{D}^{(t)}|\}\) 

        On the other hand,  the parameter $\lambda^*_i$ for the new label $\mathbf{y}_i^{*}(\lambda^*_i)$ is optimized with Eq.6 to infer the correct label for the samples, enabling uniform supervised learning. The pseudo-label \(\mathbf{y}_i^{*}\) is defined as: 
    \begin{equation*}
        \mathbf{y}_i^{*} = \begin{cases}
            \mathbf{y}_i & \text{if } {\lambda}^*_i > 0 \\
            f_{\theta}(\mathbf{x}_i) & \text{otherwise},
        \end{cases}
        \label{eq:pseudo_label}
    \end{equation*}

        %While the optimization process for \(\omega\) is still the same as in \cref{eq:traditional_update_w}, the update for \(\lambda\) can be performed as follows:
    where ${\lambda}^*_i$ is the gradient w.r.t. $\lambda_i$ of the final loss over the validation set $\mathcal{D}^{}(v)$:
    \begin{equation}
        {\lambda}^*_i =  \left [ \mathsf{sign} \left ( -\mathbb{E}_{(\mathbf{x}_{j}, \mathbf{y}_{j}) \in \mathcal{D}^{(v)}} \left[ \pdv{}{\lambda_{i}} \ell \left( f_{\theta^{*} ( \omega,\lambda )}(\mathbf{x}_{j}), \mathbf{y}_{j} \right) 
        \right ) \right] \right ]_{+},
        \label{eq:traditional_update_lambda}
    \end{equation}
    where \([\mathsf{sign}(.)]_{+}=\max(\mathsf{sign}(.),0)\) denotes the rectification operator, and \(\mathsf{sign}\) represents the sign function.

    The estimated \(\omega^{*}\), \(\hat{\mathbf{y}}_i(\lambda_i)\) and \(\mathbf{y}_i^{*}\) are then used to train a classifier of interest~\citep{Distill_noise}. Given the pseudo label \(\mathbf{y}_i^{*}\), the \emph{Distill} method further employs supervised learning with images obtained via the \emph{mixup} operator~\citep{zhang2018mixup} using the training and validation sets. Additionally, they adopt a KL-divergence loss between the model's outputs on the original and augmented inputs to enhance the consistency of the pseudo-label distribution. The final objective of \emph{Distill} framework is defined as:

    \begin{equation}
        \begin{aligned}[b]
         \mathsf{L} & = \mathbb{E}_{(\mathbf{x}_{i}, \mathbf{y}_{i}) \in \mathcal{D}^{(t)}} \left[ \omega^{*}_i \ell \left( f_{\theta}(\mathbf{x}_i), \hat{\mathbf{y}}_i(\lambda_i) \right) + \frac{\ell \left( f_{\theta}(\mathbf{x}_i), \mathbf{y}_i^{*}(\lambda^*_i)\right)}{B}  + p \, \ell \left( \mathbf{y}_i^{\beta},f_{\theta}(\mathbf{x}_i^{\beta}) \right) +  k \, \operatorname{KL} \left[ f_{\theta} \left( \mathbf{x}_i \right) \Vert f_{\theta} \left( \hat{\mathbf{x}}_i \right) \right] \vphantom{\frac{1}{1}} \right],
        \end{aligned}
        % $}
        \label{eq:training_loss_distill_supervision}
    \end{equation}

    \(\mathbf{y}_i^{\beta}\) and \(\mathbf{x}_i^{\beta}\) are obtained via the \emph{mixup} operator~\citep{zhang2018mixup} using the training and validation sets, \(\operatorname{KL}[.\Vert.]\) denotes the Kullback-Leibler (KL) divergence, \(\hat{\mathbf{x}}_i\) is an augmented sample of \(\mathbf{x}_{i}\), and \(p\) and \(k\) are hyper-parameters, \(B\) is the batch size.

\section{Methodology}
\label{sec:method}

    The meta-learning Distill model \citep{Distill_noise} presented in \cref{sec:meta_relabelling} optimises meta-parameters $\lambda$ and $\omega$ under the guidance of a validation subset \(\mathcal{D}^{(v)}\), and we adopt a similar approach for our framework. However, instead of selecting a fixed validation subset at the beginning of training as the Distill model, we propose a definition for \emph{sample informativeness} within a meta reweighting framework, as well as an iterative mechanism to automatically select a \say{high-utility} validation subset at the start of each epoch. This section presents the utility criteria to select and the mechanism to update such a validation set.
    % , using a novel utility that we will propose in detail in \cref{sec:maximising_utility_validation_set}.

    \begin{figure*}[t]
        % \vspace{-0.5em}
        \centering
        \begin{tikzpicture}[scale=0.58, every node/.style={transform shape}, font=\large]
    % setup some constants
    \pgfmathsetmacro{\xshift}{16.5}
    \pgfmathsetmacro{\yshift}{12}

    % define nodes
    \node[cylinder, shape border rotate=90, shape aspect=0.25, draw=black, fill=BurntOrange!30, minimum width=8em, text width=7em, minimum height=10em, align=center, inner sep=0.25em] at (0, 0) (training_set) {Noisy-label training set \(\mathcal{D}\)};

    \node[rectangle, draw=black, fill=violet!10, minimum width=10em, minimum height=5em, text width=10em, align=center] at ([xshift=0.75*\xshift em]training_set) (pseudo_clean_detector) {1) Pseudo-clean detector};

    \node[cylinder, shape border rotate=90, shape aspect=0.25, draw=black, fill=Red!30, minimum width=8em, text width=7em, minimum height=5em, align=center, inner sep=0.25em] at ([yshift=-\yshift em]pseudo_clean_detector) (pseudo_noisy_set) {Pseudo-noisy set \(\mathcal{D}^{(n)}\)};

    \node[cylinder, shape border rotate=90, shape aspect=0.25, draw=black, fill=ForestGreen!30, minimum width=8em, text width=7em, minimum height=7em, align=center, inner sep=0.25em] at ([xshift=\xshift em]pseudo_clean_detector) (pseudo_clean_set) {Pseudo-clean set \(\mathcal{D}^{(c)}\)};

    \node[rectangle, draw=black, fill=violet!10, minimum width=10.5em, minimum height=10em, text width=10.5em, align=center] at ([xshift=0.75*\xshift em]pseudo_clean_set) (maximise_utility) {2) Maximise utility: informativeness, cleanliness and class-balance in validation set};
    \node[rectangle, draw=none, label=above:{share classifier feature extractor}] at ([yshift=0.8125*\yshift em]maximise_utility) (shared_feature_extractor) {};

    \node[cylinder, shape border rotate=90, shape aspect=0.25, draw=black, fill=ForestGreen!30, minimum width=8em, text width=7em, minimum height=6em, align=center, inner sep=0.25em] at ([xshift=\xshift em, yshift=0.375*\yshift em]maximise_utility) (train_set) {\(\mathcal{D}^{(t)}\)};
    \node[cylinder, shape border rotate=90, shape aspect=0.25, draw=black, fill=ForestGreen!50, minimum width=8em, text width=7em, minimum height=5em, align=center, inner sep=0.25em] at ([xshift=\xshift em, yshift=-0.375*\yshift em]maximise_utility) (val_set) {\(\mathcal{D}^{(v)}\)};

    \node[rectangle, draw=black, fill=violet!10, minimum width=7em, minimum height=5em, text width=7em, align=center] at ([xshift=1.9*\xshift em]maximise_utility) (meta_learner) {3) Meta-learner};

    \node[rectangle, draw, minimum width=10em, minimum height=5em, text width=10em, align=center] at ([xshift=\xshift em, yshift=-\yshift em]maximise_utility) (labeller) {Robust labels for samples in \(\mathcal{D}^{(c)}\)};

    % connect
    \draw[-Latex] (training_set) -- (pseudo_clean_detector);
    
    \draw [-Latex] (pseudo_clean_detector) -- node[above]{pseudo-clean} node[below]{samples} (pseudo_clean_set);
    \draw[-Latex] (pseudo_clean_detector) -- node[right, text width=12em, align=center, rotate=0, anchor=north, xshift=5em, yshift=1.5em]{samples likely with noisy labels} (pseudo_noisy_set);

    \draw[-Latex] (pseudo_clean_set) -- (maximise_utility);

    \draw[-Latex] (maximise_utility.east) -- node[xshift=-0.5em, yshift=1.5em, rotate=35, anchor=north]{train set} (train_set.west);
    \draw[-Latex] (maximise_utility.east) -- node[below, rotate=-34, anchor=north]{validation set} (val_set.west);

    \draw[-Latex] (train_set.east) -- (meta_learner.west);
    \draw[-Latex] (val_set.east) -- (meta_learner.west);

    \draw[-Latex] (meta_learner.south) |- node[left=of meta_learner, pos=0.5, xshift=4em, text width=10em, align=center]{} (labeller.east);
    \draw[-Latex] (labeller.west) -| node[right=of maximise_utility, pos=0.5, xshift=-2em, text width=10em, align=center]{} (maximise_utility.south);

    \draw[densely dashed] (pseudo_clean_detector) |- (shared_feature_extractor);
    \draw[densely dashed] (shared_feature_extractor) -| (meta_learner);
    \draw[densely dashed] (shared_feature_extractor) -- (maximise_utility);
\end{tikzpicture}
        % \vspace{-0.5em}
        \caption{Main stages of INOLML: 1) classify the noisy-label samples from $\mathcal{D}$ into $\mathcal{D}^{(c)}$ (samples that are likely to have clean labels) and  $\mathcal{D}^{(n)}$ (samples likely to have noisy labels); 2) build a validation set $\mathcal{D}^{(v)}$ containing samples that are informative (from a meta-learning perspective), balanced and with a high likelihood of containing clean labels,
        %tested with the moving average robust labeller \gustavo{unclear what this means: tested with the moving average robust labeller}, 
        %where the training set $\mathcal{D}^{(t)} = \mathcal{D}^{(c)}\setminus \mathcal{D}^{(v)}$; 
        and 3) train the meta-learning classifier with $\mathcal{D}^{(t)} = \mathcal{D}^{(c)}\setminus \mathcal{D}^{(v)}$ and $\mathcal{D}^{(v)}$. 
        %\gustavo{In the diagram, the dashed line should connect 1), 2), 3), so the link to $\mathcal{D}$ should be moved to link 1) Pseudo-clean detector.}
        %\cuong{I correct it by connecting 1, 2, 3.}
        % These three steps are iterated during training.
        }
        \label{fig:motivation}
        % \vspace{-0.5em}
    \end{figure*}

    Specifically, we propose a 2-step process (corresponding to steps 1 and 2 in Figure~\ref{fig:motivation} to select a high-utility validation set to train a meta-learning model presented in \eqref{eq:distill_supervision_label_noise}. In the first step, we split the original noisy training set \(\mathcal{D}\) into pseudo-clean set \(\mathcal{D}^{(c)}\) and pseudo-noisy set \(\mathcal{D}^{(n)}\). In the second step, we further split the pseudo-clean set \(\mathcal{D}^{(c)}\) into a training subset \(\mathcal{D}^{(t)}\) and a validation subset \(\mathcal{D}^{(c)}\) based on some criteria, such as cleanliness, informativeness and class-balanced distribution.

    \subsection{Detecting noisy and clean label subsets from \texorpdfstring{$\mathcal{D}$}{D}}
    \label{sec:detecting_noisy_and_clean_label_subsets}
        Formally, the original training set is split into two subsets: clean and noisy, following a certain criterion:
        \begin{equation}
            \begin{aligned}[b]
                \mathcal{D}^{(c)} = \mathsf{PseudoCleanDetector}\left (\mathcal{D} \right ) \quad \wedge \quad \mathcal{D}^{(n)} = \mathcal{D}~\backslash~\mathcal{D}^{(c)}.
            \end{aligned}
            \label{eq:pseudo_detection}
        \end{equation}
        We first calculate the cross-entropy losses between noisy labels and the prediction \(f_{\theta}(.)\) of all samples in the training set \(\mathcal{D}\). We then apply the \emph{small loss hypothesis}~\citep{han2018co, li2020dividemix} through the notation \(\mathsf{PseudoCleanDetector}(.)\) in~\eqref{eq:pseudo_detection} to select samples having small losses to form a pseudo-clean set $\mathcal{D}^{(c)}$, because they are more likely to be clean. 
        Note that the pseudo-clean set $\mathcal{D}^{(c)}$ is independent from the target model since \(\mathcal{D}^{(c)}\) is initialised using a warm-up model (i.e., a model trained on \(\mathcal{D}\) for a few epochs), while the target model is trained from scratch. Hence, the small-loss samples in the pseudo-clean set $\mathcal{D}^{(c)}$ are likely informative to the target model. The remaining samples form the pseudo-noisy set $\mathcal{D}^{(n)}$, as shown in \eqref{eq:pseudo_detection}.
        %\gustavo{This comment is true only for the first training iteration, and the claim that the set is informative for the target set sounds too strong. I'd probably change this as follows:}
        %\gustavo{Note that the pseudo-clean set $\mathcal{D}^{(c)}$ is independent from the target model since \(\mathcal{D}^{(c)}\) is initialised using a warm-up model (i.e., a model trained on \(\mathcal{D}\) for a few epochs), while the target model is trained from scratch. Hence, the small-loss samples in the pseudo-clean set $\mathcal{D}^{(c)}$ are potentially informative to the target model, particularly at this initial training stage. The remaining samples form the pseudo-noisy set $\mathcal{D}^{(n)}$, as shown in \eqref{eq:pseudo_detection}.}
        These two sets, $\mathcal{D}^{(c)}$ and $\mathcal{D}^{(n)}$,
    are regularly updated during training.

    \subsection{Maximising the utility of the validation set}
    \label{sec:maximising_utility_validation_set}
        The pseudo-clean set \(\mathcal{D}^{(c)}\) obtained in \eqref{eq:pseudo_detection} is then divided into a validation set \(\mathcal{D}^{(v)}\) and a training set \(\mathcal{D}^{(t)}\). The validation set \(\mathcal{D}^{(v)}\) must have the following properties:
        \begin{itemize}
            \item \emph{class-balanced:} containing the same number of samples per class.
            \item \emph{informative:} being useful to guide the meta-learning framework. We define validation samples with high \emph{informativeness} as samples that can assign larger weight \(\omega_{i}\) for clean samples, and lower weight for noisy samples.
            \item \emph{clean:} most likely having clean labels.
        \end{itemize}
        To enforce such properties, we propose the objective function of the validation set as follows:
        \begin{equation}
            \begin{aligned}[b]
            & \mathcal{D}^{(t)} = \mathcal{D}^{(c)}~\backslash~\mathcal{D}^{(v)}_{\mathrm{opt}} \\
            & \mathcal{D}^{(v)}_{\mathrm{opt}}  = \operatorname*{argmax}_{\substack{\mathcal{D}^{(v)} \subset \widetilde{\mathcal{D}}^{(v)}_{\mathrm{opt}}  \\ \abs{\mathcal{D}^{(v)}} = M \times C}}
            \mathsf{Clean} \left( \mathcal{D}^{(v)},\mathcal{D}^{(c)} \right) \\
            & \text{s.t.: }
             \widetilde{\mathcal{D}}^{(v)}_{\mathrm{opt}}  = \operatorname*{argmax}_{\substack{ \widetilde{\mathcal{D}}^{(v)} \subset {\mathcal{D}}^{(c)} \\ \abs{\widetilde{\mathcal{D}}^{(v)}} = K \times C}}
             \mathsf{Info} \left( \widetilde{\mathcal{D}}^{(v)},\mathcal{D}^{(c)} \right),
            \end{aligned}
           \label{eq:validation_set_criteria}
        \end{equation}
        where \(K\) and \(M\) are the number of samples per class with \(M \le K\) to guarantee a class-balanced distribution for \(\mathcal{D}^{(v)}\), and the functions \(\mathsf{Clean}()\) and \(\mathsf{Info}()\) denote the cleanliness and informativeness of samples. $\widetilde{\mathcal{D}}^{(v)}_{\mathrm{opt}}$ and $\mathcal{D}^{(v)}_{\mathrm{opt}}$ respectively denote the optimized $\widetilde{\mathcal{D}}^{(v)}$ and $\mathcal{D}^{(v)}$.

        In other words, the lower-level in \eqref{eq:validation_set_criteria} means to determine a \say{coarse} validation set \(\widetilde{\mathcal{D}}^{(v)}\) that is class-balanced (containing \(K\) samples per class) and informative from the pseudo-clean set \(\mathcal{D}^{(c)}\). The upper-level in \eqref{eq:validation_set_criteria} refines the coarse validation set \(\widetilde{\mathcal{D}}^{(v)}\) further to form a \say{fine} subset \(\mathcal{D}^{(v)}\) containing most likely clean samples. The need for this bi-level optimization is because in practice, as the training progress, we find that informative samples need to become harder samples (samples with low confident prediction from the model) over time, hence more likely to be noisy. Note that while clean and informative samples are beneficial for the meta reweighting framework, informative-but-noisy validation samples will make the model prone to overfitting. Our optimization for the validation set \(\mathcal{D}^{(v)}\) hence aim to find the balance between class-balance, informativeness and cleanness. The details of \(\mathsf{Clean}(.)\) and \(\mathsf{Info}(.)\), are presented in the following subsubsections.

        \subsubsection{Informativeness}
            As described above, \emph{informative} samples for the validation set of a meta reweighting framework should be able to upweight clean samples, and downweight noisy samples effectively. Based on this motivation,
            our \(\mathsf{Info}(.)\) function in the lower-level of \eqref{eq:validation_set_criteria} is motivated by \emph{learning to reweight}~\citep{L2W} (similar to \eqref{eq:meta_learning_learn2reweight}). According to~\cite[Eq. (12)]{L2W}, the gradient w.r.t. \(\omega\) can be written as follows:
            \begin{equation}
                \begin{aligned}[b]
                    & \mathbb{E}_{(\mathbf{x}_{j}, \mathbf{y}_{j}) \in \mathcal{D}^{(v)}} \left[\eval{ \pdv{}{\omega_{i}} \ell \left( f_{\theta^{*}(\omega, \lambda)}(\mathbf{x}_{j}), \mathbf{y}_{j} \right) }_{\omega_i=0} \right]  \propto - \mathbb{E}_{(\mathbf{x}_{j}, \mathbf{y}_{j}) \in \mathcal{D}^{(v)}} \sum_{l=1}^{L} \left( {\mathbf{z}^{(v)}_{j,l-1}}^{\top} \mathbf{z}^{(t)}_{i,l-1} \right) \left( {\mathbf{g}^{(v)}_{j,l}}^{\top}\mathbf{g}^{(t)}_{i,l} \right),
                \end{aligned}
                \label{eq:gradient_w}
            \end{equation}
            where $\mathbf{z}^{(v)}_{j,l-1}$ is the feature of the validation sample $\mathbf{x}_{j}$ processed by a deep model at the layer $l$-th (similar definition is applied for the training feature $\mathbf{z}^{(t)}_{i,l-1}$), and $\mathbf{g}^{(v)}_{j,l}$ is the gradient from layer $l$ for the validation sample $\mathbf{x}_{j}$ (similar definition is applied for the training gradient $\mathbf{g}^{(t)}_{i,l}$).

            The re-weighting factor \(\omega_{i}\) of a training sample \(\mathbf{x}_{i}\) is, therefore, high if its feature and gradient are similar to the validation samples' feature and gradient; else, the weight is low. Hence, a validation set that maximises the weight \(\omega\) of training samples also maximises its meta-learning optimization utility. This observation is crucial to our validation sample selection.
            Intuitively, we can define the sample informativeness function $\mathsf{Info}(.)$ that we want to maximize as below:

            \begin{equation}
                \begin{aligned}[b]
                    \mathsf{Info}(\widetilde{\mathcal{D}}^{(v)},\mathcal{D}^{(c)}) = \sum_{(\mathbf{x}_i,\mathbf{y}_i) \in (\mathcal{D}^{(c)} \setminus \widetilde{\mathcal{D}}^{(v)})} \sum_{\substack{(\mathbf{x}_j,\mathbf{y}_j) \in \widetilde{\mathcal{D}}^{(v)}\\ \mathbf{y}_j = \mathbf{y}_i}} h(\mathbf{x}_i,\mathbf{x}_j),
                \end{aligned}
                \label{eq:naive_optimization}
            \end{equation}

        where \(h\) is analogous to a measure of similarity between samples:
        \begin{equation}
            h(\mathbf{x}_i,\mathbf{x}_j) = \sum_{l=1}^L (\mathbf{z}_{j,l-1}^{\top}\mathbf{z}_{i,l-1}) (\mathbf{g}_{j,l}^{\top}\mathbf{g}_{i,l}),
            \label{eq:iota}
        \end{equation}
        %\ref{eq:gradient_w},
        with $\mathbf{z}_{j,l-1}$ being the feature of \(\mathbf{x}_{j}\) at the input of layer $l$ (same for feature $\mathbf{z}_{i,l-1}$ of sample $\mathbf{x}_i$), and $\mathbf{g}_{j,l}$ being the validation gradient of layer $l$ from $\mathbf{x}_j$ (same for $\mathbf{g}_{i,l}$ from $\mathbf{x}_i$).
        Note that $h(., .)$ in~\eqref{eq:iota} is a part of the 
 weight \(\omega_{i}\)  for the training sample $(\mathbf{x}_i,\mathbf{y}_i)$ (see~\eqref{eq:gradient_w}), and $h(x_i, x_j)$ represents how much the validation sample $(\mathbf{x}_j,\mathbf{y}_j)$ can contribute to the weight of the clean training sample $(\mathbf{x}_i,\mathbf{y}_i)$, consists of a gradient matching term $\mathbf{g}_{j,l}^{\top}\mathbf{g}_{i,l}$ and a feature matching term $\mathbf{z}_{j,l-1}^{\top}\mathbf{z}_{i,l-1}$ between the two samples. Ideally, we want the weight of the clean samples remain high throughout the training process. Because $h(x_i, x_j)$ depends on samples gradient $\{\mathbf{g}_{i,l}\}_{l=1}^{L}$, $\{\mathbf{g}_{i,l}\}_{l=1}^{L}$ that may vary greatly during the training process, maximizing the function from \cref{eq:naive_optimization} may lead to a biased validation set that is optimal only for the current iteration, not throughout the training process. If a validation sample $(\mathbf{x}_j,\mathbf{y}_j)$ has high contribution to the weight of a clean training sample $(\mathbf{x}_i,\mathbf{y}_i)$, then they may share similar characteristics. Consequently, their gradient matching $\mathbf{g}_{j,l}^{\top}\mathbf{g}_{i,l}$ and feature matching $\mathbf{z}_{j,l-1}^{\top}\mathbf{z}_{i,l-1}$ will remain high and robust during the training process. Therefore, we need to modify the function $\mathsf{Info}(.)$, so that for each clean training sample $(\mathbf{x}_i,\mathbf{y}_i)$, there must be at least one validation sample $(\mathbf{x}_j,\mathbf{y}_j)$ that has high weight contribution to it. Such observation allows us to re-define the function $\mathsf{Info}(.)$ in the lower-level of~\eqref{eq:validation_set_criteria} as follows:
        \begin{equation}
            \mathsf{Info}(\widetilde{\mathcal{D}}^{(v)},\mathcal{D}^{(c)}) = \sum_{(\mathbf{x}_i,\mathbf{y}_i) \in (\mathcal{D}^{(c)} \setminus \widetilde{\mathcal{D}}^{(v)})} \max_{\substack{(\mathbf{x}_j,\mathbf{y}_j) \in \widetilde{\mathcal{D}}^{(v)}\\ \mathbf{y}_j = \mathbf{y}_i}} h(\mathbf{x}_i,\mathbf{x}_j), 
            \label{eq:informativeness}
        \end{equation}

% Intuitively, the maximization of the $\mathsf{Info}(.)$  function in the lower-level summation in~\eqref{eq:validation_set_criteria} (or the one in \cref{eq:informativeness}) means that we want each clean samples $(\mathbf{x}_i,\mathbf{y}_i)$ get upweighted by at least one clean validation samples $(\mathbf{x}_j,\mathbf{y}_j)$. If we simply maximize the sum of  weight of all samples, i.e., , all validation samples will converge to the same place.

        %Intuitively, forms a candidate balanced set $\widetilde{\mathcal{D}}^{(v)}$ by maximising the maximum \say{information content} that the pseudo-clean samples from $\mathcal{D}^{(c)} \setminus \widetilde{\mathcal{D}}^{(v)}$ can get from the samples in  $\widetilde{\mathcal{D}}^{(v)}$. 
        %The reason we maximise the maximum instead of the average \say{information content} is to guarantee that each clean training sample gets upweighted by at least one clean validation sample,  improving the diversity of the selected validation set. 
        In order to observe the effectiveness of this approach, we compare our modified $\mathsf{Info}(.)$ in \cref{eq:informativeness} with the naive utility from \cref{eq:naive_optimization}. %defined as:
        %\begin{equation}
            %\mathsf{MaxWeight}(\overline{\mathcal{D}}^{(v)},\mathcal{D}^{(c)}) = \sum_{\substack{(\mathbf{x}_i,\mathbf{y}_i) \in \mathcal{D}^{(c)}\setminus\overline{\mathcal{D}}^{(v)} \\(\mathbf{x}_j,\mathbf{y}_j) \in \overline{\mathcal{D}}^{(v)} \\ \mathbf{y}_j = \mathbf{y}_i}}  h(\mathbf{x}_i,\mathbf{x}_j). 
            %\label{eq:naive_optimization}
        %\end{equation}
\begin{figure*}[t]
    % \vspace{-0.5em}
    \centering
    \begin{subfigure}[t]{0.48 \linewidth}
        \includegraphics[width = \linewidth]{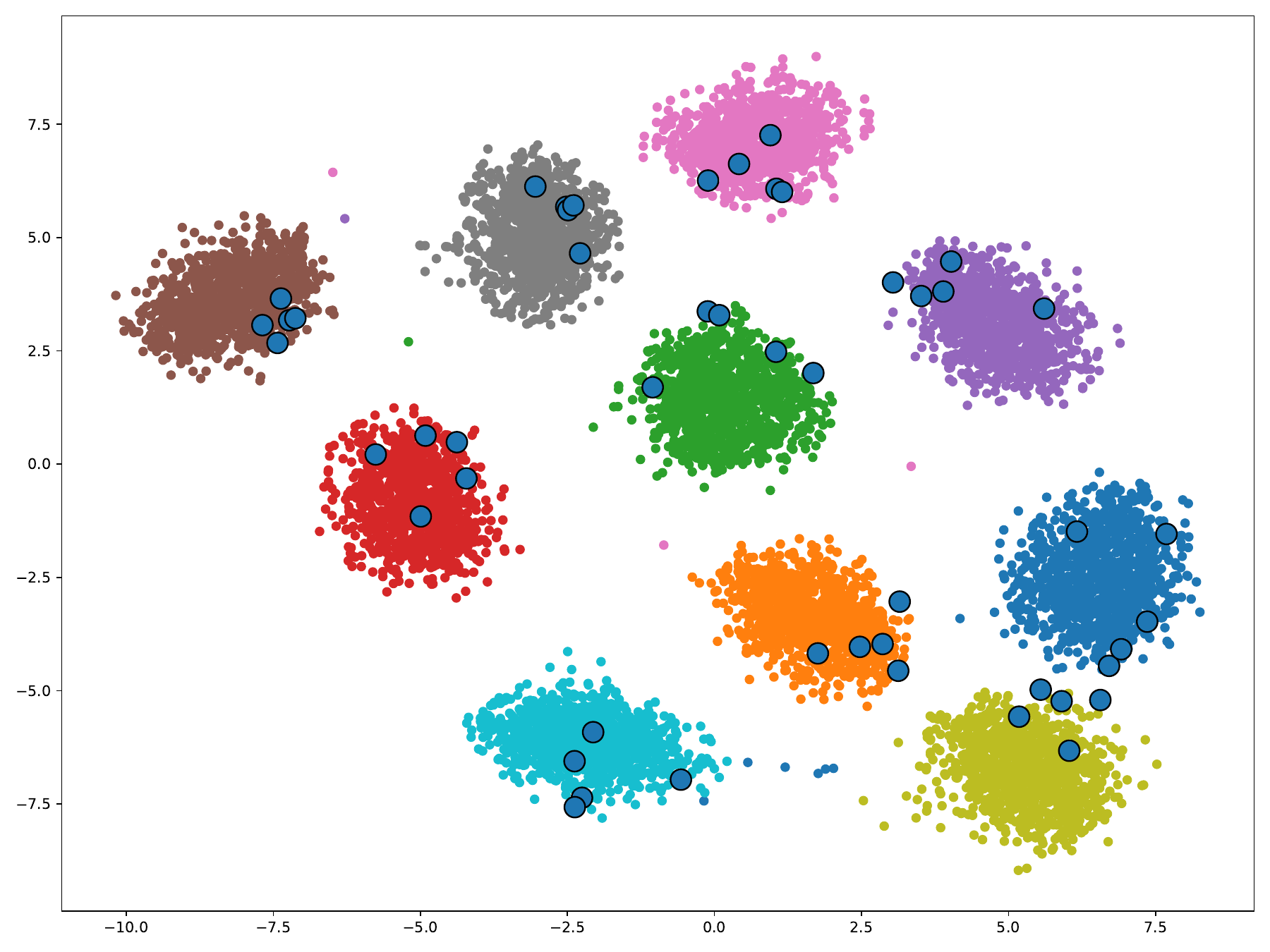}
        \caption{Naive utility (\cref{eq:naive_optimization}).}
        \label{fig:naive_optimization}
    \end{subfigure}
    \hfill
    \begin{subfigure}[t]{0.48 \linewidth}
        \includegraphics[width = \linewidth]{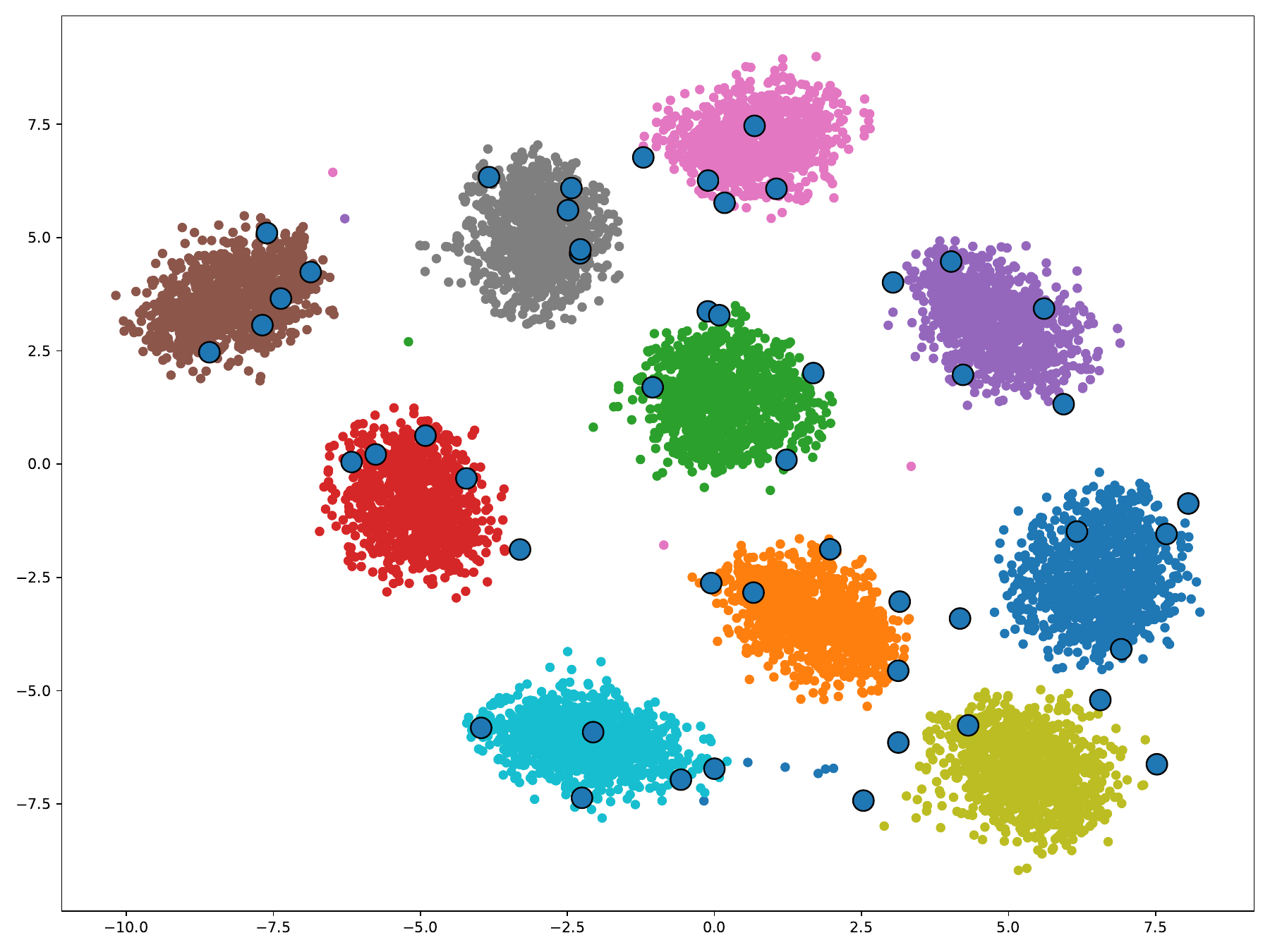}
        \caption{Our utility (\cref{eq:informativeness}).}
        \label{fig:ours}
    \end{subfigure}

    % \begin{subfigure}[t]{0.475 \linewidth}
    %     \includegraphics[width = \linewidth]{img/open_set_noise.pdf}
    %     \caption{}
    %     \label{fig:open_set_noise}
    % \end{subfigure}
    \caption{Comparison between 2-dimensional t-SNE representations of the samples selected (samples for each class have different colours, and the selected validation samples per class are highlighted with a blue dot with black outline) by (\subref{fig:naive_optimization}) naive utility in~\cref{eq:naive_optimization}, and (\subref{fig:ours}) our utility in~\cref{eq:informativeness}.  Note that this is the t-SNE representation for CIFAR10 dataset with a uniform noise rate of 40\%.}
    \label{fig:comparision_tsne}
\end{figure*}
Figure~\ref{fig:comparision_tsne} shows a visual comparison between the naive utility in~\cref{eq:naive_optimization} and our utility in~\cref{eq:informativeness} for the selection of validation samples from the pseudo-clean set $\mathcal{D}^{(c)}$ using t-SNE projection to a 2-dimensional space for CIFAR10 dataset with a uniform noise rate of 40\%.
         %\gustavo{please tell the dataset and noisy-label problem being dealt with here.}. 
         It is noticeable that the selected validation samples using our proposed utility are diverse and distributed around the clusters of their respective classes, while the validation samples selected with the naive utility in \cref{eq:naive_optimization} tend to be located closer to each other and inside each class cluster. %, so we can say that they are clean, but have lower diversity, compared with the samples formed by our utility function.
         This is because the naive utility favors low-confidence samples at the selection time. However, as training progresses and the prediction confidence of these samples increases, other samples should become the target for upweighting instead, making the selected validation set no longer optimal.
         Another observation is that our utility maximization results in a validation set that constructs a boundary between each class in feature space, helping our method to identify the classification boundary between different classes.  
        %The lower-level optimization in~\eqref{eq:validation_set_criteria} can also be explained by the fact that $\mathsf{Info(A,B)}$  expresses how informative the set $A$ is if we use the set $B$ as the validation set and conversely, making $\overline{\mathcal{D}}^{(v)}$ the most informative subset of size $K$ x $C$ we can extract from $\mathcal{D}^{(c)}$ %\dung{another explanation for lower optimization}.

        \begin{algorithm}[t]
            \caption{Training procedure of the proposed INOLML.}
            \label{algorithm:training_procedure}
            \begin{algorithmic}[1]
                \Procedure{Train}{$\mathcal{D}$, $\eta$, $T$, $\widetilde{T}$, $T^{(u)}$, $\widetilde{\eta}$, $\kappa$, $N$, $M$, $K$, $C$}
                    \LComment{\(\mathcal{D}\): noisy training set, \(\{\eta_t\}_{t}^{T}\): learning rates}
                    %\LComment{\(\eta\): learning rate per iteration}
                    \LComment{\(T\): total number of iterations}
                    \LComment{\(\widetilde{T}\): minimum number of iterations before  updating the  robust labels}
                    \LComment{\(T^{(u)}\): interval between updates}
                    %\LComment{\(\widetilde{\eta}\): learning rate threshold to update the robust labels}
                    %\LComment{\(\eta_{t}\):  learning rate at the $t^{th}$ iteration}
                    \LComment{\(\kappa,N, M, K, C\): hyper-parameters}
    
                    %\Statex
    
                    \State Warm-up $f_{\theta}(.)$ with \(\ell_{\mathrm{CE}}(.)\) on $\mathcal{D}$
                    \State $\mathcal{D}^{(c)} \gets \mathsf{PseudoCleanDetector}\left (\mathcal{D} \right )$ \Comment{\cref{eq:pseudo_detection}}
                    \State Initialise  robust label $\{\tilde{\mathbf{y}}_i\}_{i=1}^{|\mathcal{D}^{(c)}|}$ of samples in $\mathcal{D}^{(c)}$
                    \State Initialise $\mathcal{D}^{(v)}$ and $\mathcal{D}^{(t)}$ from $\mathcal{D}^{(c)}$ \Comment{\cref{eq:validation_set_criteria}}
                    \State Reinitialize $f_{\theta}(.)$
                    \For{$t = 1$ to $T$}
                        \State Meta-learn \(\theta\), \(\omega\) and \(\lambda\) \Comment{\cref{eq:distill_supervision_label_noise}}
                        \If{($ t > \widetilde{T}$)}% and  ($\eta_t < \widetilde{\eta}$)}
                            \State Update  $\{\tilde{\mathbf{y}}_i\}_{i=1}^{|\mathcal{D}^{(c)}|}$ of samples in $\mathcal{D}^{(c)}$ \Comment{\cref{eq:moving_average}}
                        \EndIf
                        %\If{$\eta_t = 0$}
                        %\State  %$\mathcal{D}^{(c)}=\mathsf{PseudoCleanDetector}\left (\mathcal{D} \right )$ \Comment{\cref{eq:pseudo_detection}}
                        %\EndIf
                        \If{$t \mod T^{(u)}=0$}
                            \State  Construct $\widehat{\mathcal{D}}^{(c)}$. \Comment{\cref{eq:refine_pseudo_set}}
                            \State Select new $\mathcal{D}^{(v)}$  from $\widehat{\mathcal{D}}^{(c)}$ \Comment{\cref{eq:validation_set_criteria}}
                        \EndIf
                    \EndFor
                    \State \Return the trained model parameter \(\theta\)
                \EndProcedure
            \end{algorithmic}
        \end{algorithm}
        
        \subsubsection{Cleanliness}
        
        % from the  subset $\overline{\mathcal{D}}^{(v)}$ of informative candidate samples , we want to select the most confident samples for  the final validation set $D^{(v)}$. \dung{emphase informativeness is more important than cleanness}
        
        %Simply filtering out samples with large gradients will produce a validation set with clean but less informative samples.
        Note that our focus on the informativeness of the validation set should not compromise the cleanliness of samples, as meta-learning under the guidance of informative-but-noisy validation samples will make the model prone to overfitting. Indeed, the samples in the \say{coarse}-but-informative validation set $\widetilde{\mathcal{D}}^{(v)}$ can still have noisy labels since $\mathcal{D}^{(c)}$ is not completely clean. In addition, the function \(\mathsf{Info}(x_i,x_j)\) prioritizes samples in $\widetilde{\mathcal{D}}^{(v)}$ with large gradient magnitude, as it has a gradient matching component $\mathbf{g}_{j,l}^{\top}\mathbf{g}_{i,l}$. Consequently, the selected validation samples are usually \say{hard samples} and more likely to be noisy. It is, therefore, important to balance the trade-off between sample informativeness and cleanness. However, as gradient magnitude is one of our main factors to identify informative samples, we do not want to minimize their gradient magnitude for cleanliness metric. By observing that the samples from  \(\tilde{\mathcal{D}}^{(v)}\) are more likely to be clean when they have high similarity with other samples in the same class, we  propose a heuristics to identify the samples of interest with high chance being clean. Specifically, we employ the cosine similarity between the samples of interest  and other samples in the same class in \(\mathcal{D}^{(c)}\), defined as follows:
        \begin{equation}
        % \scalebox{0.9}{$
            \mathsf{Clean} \left( \mathcal{D}^{(v)},\mathcal{D}^{(c)} \right) =  \sum_{\substack{(\mathbf{x}_j,\mathbf{y}_j) \in \mathcal{D}^{(v)}\\
            (\mathbf{x}_i,\mathbf{y}_i) \in \mathcal{D}^{(c)}\setminus\mathcal{D}^{(v)} \\ \mathbf{y}_i = \mathbf{y}_j}}
             \sum_{l=1}^L \left(\mathbf{z}_{j,l-1}^{\top}\mathbf{z}_{i,l-1} \right).
             % $}
             \label{eq:clean}
        \end{equation}
        % To guarantee a class-balanced distribution for \(\mathcal{D}^{(v)}\), we select \(M\) samples for each of the \(C\) classes with \(M \ll K\), as shown in the upper-level of~\eqref{eq:validation_set_criteria}. % to obtain a balanced  \(\mathcal{D}^{(v)}\).
        
        Both combinatorial optimizations in~\eqref{eq:validation_set_criteria}, and in particular \cref{eq:informativeness,eq:clean}, are solved with greedy strategies. We initially loop through each class and select \(K\) samples per class to maximise the lower-level objective function. We then sequentially select \(M\) samples among the previous set of \(K\) samples per each class to optimise the upper-level objective function.
        %\dung{I explained how the greedy work here}. 
        We simplify the calculation of gradient in~\eqref{eq:gradient_w} and the optimization in~\eqref{eq:validation_set_criteria} by using only the features and gradients in the last layer of the deep neural network of interest. This simplification is reasonable since according to~\citep{FSR}, the weights of training samples in meta-learning  depend mostly on the last layer of the  model, and for datasets with a large number of classes, the memory complexity is intractable. %\gustavo{add explanation from the rebuttal here.}
        For instance, we need around $K=200$ candidate samples per class for $\tilde{\mathcal{D}}^{(v)}$ before selecting $M=10$ samples per class for the validation set $\mathcal{D}^{(v)}$ (e.g., for CIFAR100, we need 20,000 samples for $\tilde{\mathcal{D}}^{(v)}$), where for each sample, we have to store its gradient and feature embedding for every layer. For example, for Resnet18 that has $1\times10^7$ parameters, if we needed to store all features and gradients from all layers for the whole $\tilde{\mathcal{D}}^{(v)}$, we would need to store $(2 \times 10^4) \times (1.1 \times 10^7)$ floating point numbers (i.e., $\approx 10^{12}$ bytes), which is intractable.
        
        %\footnote{For CIFAR100, we  need to compute and store gradients and feature embeddings for 20000 samples, assuming that we require 200 candidate samples per class. Hence, for Resnet18 that has 11 million parameters, we need to store $(2 \times 10^4) \times (1.1 \times 10^7)$ floating point numbers for gradients and embeddings, which is around $10^{12}$ bytes of memory (i.e., 1 TB).}.
    \subsection{Dynamic Pseudo Clean Set Refinement}
    To further refine the pseudo clean set $\mathcal{D}^{(c)}$, we adopt the robust moving average label of each sample which is defined as follows:
    \begin{equation}
        \tilde{\mathbf{y}}_i = \kappa \tilde{\mathbf{y}}_i +  (1-\kappa)\frac{1}{E} \sum_{e=1}^{E} f_{\theta}(\mathbf{x}_i),
    \label{eq:moving_average}
    \end{equation}
    with \(\kappa = 0.9\).
   The robust label is updated regularly, and becomes more accurate as the model improves. At the start of each data selection step, we eliminate potentially noisy samples from the pseudo clean set that were previously overlooked by the $\mathsf{PseudoCleanDetector}$ using the following method:
    \begin{equation}
    \widehat{\mathcal{D}}^{(c)} = \left\{ \vphantom{\max_{k\in\{1,...,C\}}} (\mathbf{x}_{i}, \mathbf{y}_{i}): (\mathbf{x}_{i}, \mathbf{y}_{i}) \in \mathcal{D}^{(c)}  \wedge \, \operatorname*{argmax}_{k\in\{1,...,C\}} \mathbf{y}_{i}(k) = \operatorname*{argmax}_{k\in\{1,...,C\}} \tilde{\mathbf{y}}_{i}(k) \right\}.
    \label{eq:refine_pseudo_set}
    \end{equation}
    We utilize the proxy set $\widehat{\mathcal{D}}^{(c)}$ for our data selection algorithm instead of the full set $\mathcal{D}^{(c)}$.
    
    \subsection{Training procedure}
    \label{sec:training_procedure}
    
    Our training follows the 3-step iterative approach in Figure~\ref{fig:motivation}, where step 1 (pseudo-clean label detector) and step 2 (utility maximisation of the validation set) have been explained above, and
    step 3 (meta-learning) is based on~\eqref{eq:distill_supervision_label_noise} with the loss $\mathsf{L}$ defined in \cref{eq:training_loss_distill_supervision}.
    The details of the training process are shown in \cref{algorithm:training_procedure}.

    \section{Experiments}
\label{sec:experiments_and_analysis}

\begin{table*}[t!]
    \centering
    \caption{Test accuracy (\%) of INOLML and previous methods for symmetric noise; methods with \textsuperscript{T} represent meta-learning methods that need clean validation sets; the lower block contains meta-learning methods, while the upper block shows SOTA methods.}
    \label{tab:table_2}
    % \fontsize{13}{9}
    % \scalebox{0.9}{
    \begin{tabular}{l c c c c c c}
    \toprule
    \multirow{2}{*}{\bfseries Method} %\multicolumn{6}{c}{\bfseries Dataset} \\
    % \cmidrule{2-7}
    & \multicolumn{3}{c}{\bfseries CIFAR10} &  \multicolumn{3}{c}{\bfseries CIFAR100}  \\
    \cmidrule(lr){2-4} \cmidrule(lr){5-7}
    & \multicolumn{1}{c}{\bfseries 0.2} & \multicolumn{1}{c}{\bfseries 0.4} & \multicolumn{1}{c}{\bfseries 0.8} & \multicolumn{1}{c}{\bfseries 0.2} & \multicolumn{1}{c}{\bfseries 0.4} & \multicolumn{1}{c}{\bfseries 0.8} \\ 
    \midrule
    %GCE\citep{GCE}   & 89.9 \(\pm\) 0.2 & 87.1 \(\pm\) 0.2 & 67.9 \(\pm\) 0.6 & 66.8 \(\pm\) 0.4 & 61.8 \(\pm\) 0.2 & 47.7 \(\pm\) 0.7  \\
    GJS & 95.3 \(\pm\) 0.2 & 93.6 \(\pm\) 0.2 & 79.1 \(\pm\) 0.3 & 78.1 \(\pm\) 0.3 & 75.7 \(\pm\) 0.3 & 44.5 \(\pm\) 0.5  \\
    DivideMix   & 95.7 \(\pm\) 0.0 & - & 92.9 \(\pm\) 0.0 & 76.9 \(\pm\) 0.0 & -  & 59.6 \(\pm\) 0.0 \\
    CRUST   & 91.1 \(\pm\) 0.2 & 89.2 \(\pm\) 0.2& 58.3 \(\pm\) 1.8& - & - & - \\
    PENCIL   & - & - & - &73.9 \(\pm\) 0.3 & 69.1 \(\pm\) 0.6 & - \\
    UNICON   & 96.0 \(\pm\) 0.0 & - & 93.9 \(\pm\) 0.0 & 78.9 \(\pm\) 0.0 & - & 63.9 \(\pm\)  0.0 \\
    %ELR   & 92.1 \(\pm\) 0.4 &91.4 \(\pm\) 0.2& 80.7 \(\pm\) 0.6 & 74.7 \(\pm\) 0.3 & 68.4 \(\pm\) 0.4 & 30.2 \(\pm\)  0.8 \\
    %CausalNL   & 79.9 \(\pm\) 0.0 & - & 17.0 \(\pm\) 0.0 & - & - & - \\
    
    CausalNL + NPC   & 81.2 \(\pm\) 0.0 & - & 18.8 \(\pm\) 0.0 & - & - & - \\
    DMLP\textsuperscript{T} & 94.2 \(\pm\) 0.0 & - & 93.2 \(\pm\) 0.0  & 72.3 \(\pm\) 0.0 & -  & 63.2 \(\pm\) 0.0  \\
    DMLP-DivideMix\textsuperscript{T} & 96.2 \(\pm\) 0.0 & - & 94.3 \(\pm\) 0.0  & 79.4 \(\pm\) 0.0 & -  & 68.5 \(\pm\) 0.0 \\
    TLC & 95.0 \(\pm\) 0.1 & - & 92.5 \(\pm\) 0.2  & 78.0 \(\pm\) 0.2 & -  & 65.0 \(\pm\) 0.3 \\
     PSDC   & 96.2 \(\pm\) 0.0 & - & 94.0 \(\pm\) 0.0 & 79.4 \(\pm\) 0.0 & - & 64.3 \(\pm\)  0.0 \\
    Bayesian DivideMix++   & 96.13 \(\pm\) 0.07 & - &94.97 \(\pm\) 0.02 & 80.02 \(\pm\) 0.03 & - & 70.01 \(\pm\)  0.23 \\

    \midrule
    
    Distill\textsuperscript{T} & 96.2 \(\pm\) 0.2 & 95.9 \(\pm\) 0.2 & 93.7 \(\pm\) 0.5 & 81.2 \(\pm\) 0.7 & 80.2 \(\pm\)  0.3 & \(\mathbf{75.5} \pm \mathbf{0.2}\)\\
    MentorNet\textsuperscript{T}   & 92.0 \(\pm\) 0.0 & 89.0 \(\pm\) 0.0 & 49.0 \(\pm\) 0.0 & 73.0 \(\pm\) 0.0 & 68.0 \(\pm\) 0.0 & 35.0 \(\pm\)  0.0 \\
    L2R\textsuperscript{T}   & 90.0 \(\pm\) 0.4 & 86.9 \(\pm\) 0.2 & 73.0 \(\pm\) 0.8 & 67.1 \(\pm\) 0.1 & 61.3 \(\pm\) 2.0 & 35.1 \(\pm\)  1.2 \\
    MWN\textsuperscript{T}   & 90.3 \(\pm\) 0.6 &87.5 \(\pm\) 0.2& - & 64.2 \(\pm\) 0.3&58.6 \(\pm\) 0.5& - \\
    GDW\textsuperscript{T}  & - & 88.1 \(\pm\) 0.4 & - & - & 59.8 \(\pm\)1.6 & - \\
    
    FaMUS   & - & 95.3 \(\pm\) 0.2 & -  & - & 76.0 \(\pm\) 0.2 & -  \\
    FSR   & 95.1 \(\pm\) 0.1 &93.7 \(\pm\) 0.1 &82.8 \(\pm\) 0.3 & 78.7 \(\pm\) 0.2 & 74.2 \(\pm\) 0.4 & 46.7 \(\pm\)  0.8 \\
    \midrule
    \textbf{INOLML} & \(\mathbf{96.9}\pm\mathbf{0.1}\) & \(\mathbf{96.6}\pm\mathbf{0.1}\) & \(\mathbf{95.0}\pm\mathbf{0.2}\) & \(\mathbf{82.0}\pm\mathbf{0.2}\) & \(\mathbf{81.3}\pm\mathbf{0.2}\) & 74.7\(\pm\)0.1 \\
    \bottomrule
    \end{tabular}
    % }
\end{table*}

\subsection{Datasets} The proposed method INOLML is evaluated on several datasets, including CIFAR10, CIFAR100, mini-WebVision and Controlled Noisy Web Labels (CNWL). Both CIFAR10 and CIFAR100 datasets~\citep{Cifar_dataset} contain 50k and 10k images used for training and testing, respectively. Each image has a size of 32\(\times\)32 pixels and is labelled as one of 10 or 100 classes. 
WebVision~\citep{Webvision} is a dataset of 2.4 million images crawled from Google and Flickr based on the 1,000 ImageNet classes~\citep{deng2009imagenet}. The dataset is more challenging than CIFAR since it is class-imbalanced and contains real-world noisy labels. Following~\citep{FSR}, we extract a subset that contains the first 50 classes to create the mini-WebVision dataset~\citep{MentorNet}. 
CNWL~\citep{jiang2020beyond} is a benchmark of controlled real-world label noise that contains noise rates from 0 to 0.8. Following recent studies~\citep{Famus}, we evaluate the proposed method on the Red mini-ImageNet dataset consisting of 50k training images from 100 classes for training and 5k images for testing. Note that we use the image size of 32\(\times\)32 pixels for a fair comparison with FAMUS~\citep{Famus} and other related methods~\citep{Famus}.

% \subsection{Implementation Details}
\subsection{Implementation details}
For all experiments on CIFAR datasets, except long-tail imbalance, we use the same hyperparameters and network architectures as the Distill model~\citep{Distill_noise}.
We adopt the cosine learning rate decay with warm restarting~\citep{Cosine_lr} and SGD optimiser. For CIFAR, we train WideResnet28-10 with 100k iterations and a batch size of 100. We also train a smaller network (Resnet29) to fairly compare with~\citep{Distill_noise}. For mini-WebVision, we follow FSR~\citep{FSR} and train a single Resnet50 network 
with 1 million iterations and a batch size of 16. For Red mini-ImageNet, we run experiments with 150k iterations and a batch size of 100.
For CNWL, we use a single PreAct Resnet18 network to be consistent with previous works~\citep{propmix,ortego2021multi} on this benchmark. For the class imbalance problems, we use Resnet32 to fairly compare with FaMUS~\citep{Famus} and FSR~\citep{FSR}. We report the prediction accuracy of each experiment on their corresponding testing sets. We compare our method with recently published state-of-the-art (SOTA) meta-learning approaches, including FaMUS~\citep{Famus}, FSR~\citep{FSR}, Meta Weight Net~\citep{MetaWeightNetLA}, Distill~\citep{Distill_noise}, GDW~\citep{GDW}, L2R\citep{L2W},MSLG\citep{MetaSL}. Furthermore, some SOTA noisy labels learning approaches are also being compared with, such as DivideMix~\citep{li2020dividemix}, CausalNL~\citep{causalnl}, NPC~\citep{NPC},  MentorMix~\citep{jiang2020beyond}, MentorNet~\citep{MentorNet},  MOIT~\citep{ortego2021multi},  GJS~\citep{GJS}, CRUST~\citep{CRUST},  Co-teaching~\citep{han2018co},  Iterative-CV~\citep{Chen2019UnderstandingAU}, HAR~\citep{Imbalance_noisy},  UNICON~\citep{unicon},  NCR~\citep{Iscen2022LearningWN}, C2D~\citep{zheltonozhskii2021contrast}, BtR~\citep{Smart2022BootstrappingTR}, CC~\citep{Zhao2022CentralityAC}, SSR~\citep{Feng2021SSRAE}, DMLP~\citep{DMLP}, Twin Contrast~\citep{twin_contrast}. For imbalance learning mixed with noisy label experiments, we compare with recent noisy-label imbalanced learning methods, including ROLT~\citep{ROLT}, FSR~\citep{FSR}, LDAM~\citep{LDAM}, BBN~\citep{BBN},  HAR~\citep{Imbalance_noisy}, and CRUST~\citep{CRUST}. The supplementary material shows more implementation details.

\subsection{Symmetric noise} 

\cref{tab:table_2} shows the test accuracy results on symmetric noise rates varying from 20\% to 80\% for different meta-learning approaches that require a clean validation set (indicated with \textsuperscript{T}) and others that automatically build validation sets. 
INOLML outperforms all previous methods in most cases. The slightly lower performance than Distill on CIFAR100 at 80\% noise rate can be explained by Distill's large manually-curated clean validation set with 10 clean samples per class. In addition, as shown in Figure~\ref{fig:probe_acc}, at 80\% symmetric noise rate, a significant proportion (20\% to 45\%) of our clean validation set $\mathcal{D}^{(v)}$ contains noisy samples at the final training stages. This, however, deteriorates the efficacy of our approach.
Note that it is reasonable that the noise rate increases in the validation set as training progresses because our method prioritises the more informative samples (or harder samples) over clean ones to be included in the validation set at each epoch. Such prioritisation results in a decrease in the cleanliness of our validation set as the model gets more accurate. 
We also carry out additional experiments with different validation set sizes to fairly compare with Distill in \cref{sec:additional_symmetric_noise_results}. In particular, our method outperforms Distill by 1\% to 3\% in most scenarios.
Overall, these results show that a pseudo-clean, balanced, and informative validation set, can outperform a randomly-collected clean validation set in symmetric noise scenarios. Our results also set new SOTA results on the symmetric label noise benchmarks for methods without a clean validation set.

        \begin{figure*}[t]
           \centering
           \hspace{-1em}
            \begin{subfigure}[b]{0.49\linewidth}
               \begin{tikzpicture}
               \pgfplotstableread[col sep=comma, header=true]{latex/open_uniform_noise.txt} \myTable
               \begin{axis}[
                   height = 0.5 \linewidth,
                   width = 0.75\linewidth,
                   title style={font=\scriptsize},
                   xlabel={Iteration (\(\times\)10,000)},
                   xlabel style={font=\scriptsize, yshift=0.5em},
                   xticklabel style = {font=\scriptsize},
                   ylabel={Accuracy on $D^{(v)}$},
                   ylabel style={font=\scriptsize, yshift=-0.5em},
                   yticklabel style = {font=\scriptsize},
                   legend entries={Noise rate = 0.2, Noise rate = 0.4, Noise rate = 0.8},
                   legend style={draw=none, font=\scriptsize, legend columns=1, yshift=-3em},
                   legend image post style={scale=1},
                   legend cell align={left},
                   legend pos=north east,
                    % restrict x to domain=0:300,
                   scale only axis
               ]
                \addplot[mark=none, RoyalBlue, thick] table[x expr=0.0001*\thisrow{iteration_uniform_0.2}, y={noise_ratio_uniform_0.2}]{\myTable};
                \addplot[mark=none, OrangeRed, thick, dashed] table[x expr=0.0001*\thisrow{iteration_uniform_0.4}, y={noise_ratio_uniform_0.4}]{\myTable};
                \addplot[mark=none, ForestGreen, thick, dashdotted] table[x expr=0.0001*\thisrow{iteration_uniform_0.8}, y={noise_ratio_uniform_0.8}]{\myTable};
               \end{axis}
               \end{tikzpicture}
               \caption{Symmetric noise}
               \label{fig:probe_acc}
            \end{subfigure}
            \hfill
            \begin{subfigure}[b]{0.49\linewidth}
               \begin{tikzpicture}
               \pgfplotstableread[col sep=comma, header=true]{latex/open_noise_only.csv} \myTable
               \begin{axis}[
                   height = 0.5\linewidth,
                   width = 0.75\linewidth,
                   title style={font=\scriptsize},
                   xlabel={Iteration (\(\times\)10,000)},
                   xlabel style={font=\scriptsize, yshift=0.5em},
                   xticklabel style = {font=\scriptsize},
                   ylabel={Accuracy on $D^{(v)}$},
                   ylabel style={font=\scriptsize, yshift=-0.5em},
                   yticklabel style = {font=\scriptsize},
                   legend entries={ImgNet, Cifar100, Both},
                   legend style={draw=none, font=\scriptsize, legend columns=1},
                   legend image post style={scale=1},
                   legend cell align={left},
                   legend pos=north east,
                    % restrict x to domain=0:300,
                   scale only axis
               ]
                \addplot[mark=none, RoyalBlue, thick] table[x expr=0.0001*\thisrow{iteration_open_set_imagenet}, y={noise_ratio_open_set_imagenet}]{\myTable};
                \addplot[mark=none, OrangeRed, thick, densely dashed] table[x expr=0.0001*\thisrow{iteration_open_set_cifar100}, y={noise_ratio_open_set_cifar100}]{\myTable};
                \addplot[mark=none, ForestGreen, thick, dashdotted] table[x expr=0.0001*\thisrow{iteration_open_set_cifar100_imagenet}, y={noise_ratio_open_set_cifar100_imagenet}]{\myTable};
               \end{axis}
               \end{tikzpicture}
               \caption{Open-set noise}
               \label{fig:probe_acc_open}
            \end{subfigure}
            \caption{Accuracy of the clean validation set \(\mathcal{D}^{(v)}\) as training progresses evaluated on different noise benchmarks.}
            \label{fig:accuracy_probe_openset}
        \end{figure*}
%\vspace{-1em}
        \subsection{Asymmetric noise}
            
            \cref{tab:assymmetric_noise_distill} shows a comparison between INOLML and Distill~\citep{Distill_noise} 
            with different validation set sizes: 1, 5 and 10 samples per class.
            Although our proposed method does not rely on a manually-labelled validation set, it performs better than Distill, especially with small network architectures (Resnet29) and small validation set sizes  (1 sample per class).
            Our method has slightly lower accuracy than Distill with larger clean validation set sizes (at least 5 random clean samples per classes) on larger network architectures (WideResnet28). This might be caused by the confirmation bias of asymmetric noise in our pseudo-clean validation subset and the high capacity of larger models, e.g., WideResnet28-10, which are more prone to overfit label noise when trained on a small dataset, like CIFAR10. In addition,
            INOLML also shows better performance than SOTA methods, like FSR, DivideMix and UNICON (see \cref{tab:assymmetric_noise_SOTA}).

            %the results shows that our method outperform the Distill noise model when at least 1 of 2 conditions are available: either  the clean subset size is small to medium (1-5 sample per classes) or the model has low capacity (Resnet29 model) since bigger models are more likely to be overfitted with noise. For experiments with bigger model (WideResnet28-10) with large clean subset (10 samples per classes), our result achieve competitive but slightly less than the Distill noise model with completely clean validation set. 
            \begin{table}[t]
                \centering
                \caption{Test accuracy (\%) of INOLML on CIFAR10 with 0.4 asymmetric noise, in comparison with Distill using a validation set $\mathcal{D}^{(v)}$ of sizes 1, 5 and 10 samples per class on Resnet29 and WideResnet28-10. The superscript \textsuperscript{T} indicates the need for clean validation sets.} \label{tab:assymmetric_noise_distill}
                
                \centering
                \scalebox{1.0}{
                  \begin{tabular}{l c c c}
                        \toprule
                        \bfseries Method & \bfseries Size of validation set \(\abs{\mathcal{D}^{(v)}}\) & \bfseries Resnet29 & \bfseries WRN28-10  \\
                        \midrule
                        Distill\textsuperscript{T} & \multirow{2}{*}{$1\times C$} & 76.8 \(\pm\) 2.9 & 93.2 \(\pm\) 0.2  \\
                        \textbf{INOLML} & & 86.8 \(\pm\) 0.9 & 93.6 \(\pm\) 0.3 \\
                        \midrule
                        Distill\textsuperscript{T} & \multirow{2}{*}{$5\times C$} & 86.7 \(\pm\) 0.5 & 94.5 \(\pm\) 0.2 \\
                        \textbf{INOLML} & & 89.3 \(\pm\) 0.2 & 94.1 \(\pm\) 0.1 \\
                        \midrule
                        Distill\textsuperscript{T} & \multirow{2}{*}{$10\times C$} & 88.6 \(\pm\) 0.7 & \(\mathbf{94.9} \pm \mathbf{0.1}\)\\
                        \textbf{INOLML} & & \(\mathbf{89.8} \pm \mathbf{0.3}\) & 94.2 \(\pm\) 0.1\\
                        \bottomrule
                    \end{tabular}
                    }

            \end{table}

            \begin{table}[t]
                \centering
                \caption{Test accuracy (\%) of INOLML and previous methods on CIFAR10 with 0.4 asymmetric noise (meta-learning approaches are in the bottom part of the table). The superscript \textsuperscript{T} indicates the need for clean validation sets.} \label{tab:assymmetric_noise_SOTA}
                
                \centering
                \scalebox{1.0}{
                  \begin{tabular}{l c}
                        \toprule
                        \bfseries Method & \bfseries Accuracy \\
                        \midrule
                        %GCE\citep{GCE} & 82.3 \(\pm\) 0.7\\
                        GJS & 89.7 \(\pm\) 0.4\\
                        F-Correction   & 83.6 \(\pm\) 0.3\\
        UNICON  & 94.1 \(\pm\) 0.0\\
                        PENCIL & 91.2 \(\pm\) 0.0\\
                        DivideMix   & 92.1 \(\pm\) 0.0\\
                        %MLNT   & 92.3 \(\pm\) 0.1 \\
                        CausalNL   & 74.8 \(\pm\) 0.0 \\
                        TLC   & 92.6 \(\pm\) 0.0 \\
                        TLC+   & 93.7 \(\pm\) 0.0 \\
                        PSDC   & \(94.2 \pm 0.0\) \\
                        \textbf{DMLP\textsuperscript{T}}   & \(\textbf{95.0} \pm \mathbf{0.0}\) \\
                        
                        \midrule
                        L2R\textsuperscript{T} & 90.8 \(\pm\) 0.3\\
                        FSR & 93.6 \(\pm\) 0.3\\
                        \textbf{INOLML} & \(\mathbf{94.2} \pm \mathbf{0.1}\) \\
                        \bottomrule
                    \end{tabular}
                    }
            \end{table}
            %\vspace{-2em}

            %\begin{table}[t]
            %    \centering
            %    \caption{Comparison of our methods with FSR model and other meta-
            %    reweighting models on CIFAR10 dataset with an asymmetric noise of 0.4. Methods with \textsuperscript{T} are those require extra clean subset. The lower block are meta-learning methods while the upper block has other leading methods’s result.}
            %    \label{tab:table_4}
            %    \begin{tabular}{l c}
            %        \toprule
            %        \bfseries Method & \bfseries Accuracy \\
            %        \midrule
            %        GCE\citep{GCE} & 82.3 \(\pm\) 0.7\\
            %        F-Correction & 83.6 \(\pm\) 0.3\\
            %        PENCIL\citep{PENCIL} & 91.2 \(\pm\) 0.0\\
            %        \midrule
            %        L2R\textsuperscript{T}\citep{L2W} & 90.8 \(\pm\) 0.3\\
            %        FSR\citep{FSR} & 93.6 \(\pm\) 0.3\\
            %        \rowcolor{gray!30} \textbf{Ours} & \textbf{94.2 \(\pm\) 0.1} \\
            %        \bottomrule
            %    \end{tabular}
            %\end{table}

        %\vspace{-1em}
        \subsection{Imbalanced learning}
            %Long-tailed imbalance is another problem about the imbalance in the number of training samples  between different classes that is often studied separately from noisy label issue. 
            We evaluate INOLML on the imbalanced CIFAR datasets following  the setting in~\citep{FSR}. 
            The prediction accuracy in \cref{tab:table_6} shows that INOLML considerably surpasses other meta-learning approaches.  
            
            \begin{table}[t]
                \centering
                \caption{Test accuracy (\%) of INOLML and other SOTA meta-learning approaches evaluated on the CIFAR imbalanced learning (long-tailed) recognition task. The reported results are from~\citep{FSR} and \citep{Famus}.
                % All results are averaged over three runs.
                }
                \label{tab:table_6}
                % \scalebox{0.8}{
                \begin{tabular}{l c c c c c c}
                    \toprule
                    & \multicolumn{3}{c}{\bfseries CIFAR10} &  \multicolumn{3}{c}{\bfseries CIFAR100} \\
                    \cmidrule(lr){2-4} \cmidrule(lr){5-7}
                    \bfseries Imbalance ratio & \bfseries 200 & \bfseries 50 & \bfseries 10 & \bfseries 200 & \bfseries 50 & \bfseries 10  \\
                    % \cmidrule(lr){2-4} \cmidrule(lr){5-7}
                    \midrule
                    Softmax  &  65.68 & 74.81 & 86.39 &  34.84 & 43.85 & 55.71 \\
                    CB-Focal  &  65.29 & 76.71 & 86.66 & 32.62 & 44.32 & 55.78 \\
                    CB-Best  &  68.89 & 79.27 & 87.49  & 36.23 & 45.32 & 57.99 \\
                    \midrule
                    L2R  &  66.51 & 78.93 & 85.19 &  33.38 & 44.44 & 53.73  \\
                    MWN  &  68.91 & 80.06 & 87.84 & 37.91 & 46.74 & 58.46  \\
                    %CE+FaMUS \citep{Famus} &  - & 83.15 & 89.39 & - &  49.56 & 60.42  \\
                    %LDAM+FaMUS \citep{Famus} &  - & 83.32 & 87.90 & - &  49.93 & 59.03  \\
                    GDW  &  - & - & 86.8 & - & - & 56.8  \\
                    FaMUS  &  - & 83.32 & 87.90 & - &  49.93 & 59.03  \\
                    FSR-DF  &      66.15 & 79.78 & 88.15 & 36.74 & 44.43 & 55.60  \\
                    FSR  &  67.76 & 79.17 & 87.40 & 35.44 & 42.57 & 55.45  \\
                    \textbf{INOLML} & \(\mathbf{74.91}\) & \(\mathbf{84.43}\) & \(\mathbf{90.81}\)  & \textbf{41.52} & \textbf{51.35} & \textbf{62.07} \\
                    \bottomrule
                \end{tabular}
               % }
            \end{table}
            
            %\begin{table}[t]
            %    \centering
            %    \caption{Test accuracy (\%) comparison of our method and other leading meta-learning methods on CIFAR imbalanced learning (long-tailed) recognition. Results of other models are collected from Zhang et al.~\citep{FSR} and Xu et al.~\citep{Famus}. All results are averaged over three runs.}
            %    \label{tab:table_6}
            %    \begin{tabular}{l l l l  l l l}
            %        \toprule
            %        \bfseries Method & \multicolumn{3}{c}{\bfseries CIFAR10} &  \multicolumn{3}{c}{\bfseries CIFAR100}  \\
            %        \cmidrule{1-7} 
                    
            %        Imb. Ratio & 200 &  50 & 10 &  200 & 50 & 10  \\
            %        \cmidrule{2-4} \cmidrule{5-7}
            %        Softmax \citep{FSR} &  65.68 & 74.81 & 86.39  & 34.84 & 43.85 & 55.71 \\
            %        CB-Focal \citep{FSR} &  65.29 & 76.71 & 86.66  & 32.62 & 44.32 & 55.78 \\
            %        CB-Best \citep{FSR} &  68.89 & 79.27 & 87.49  & 36.23 & 45.32 & 57.99 \\
            %        \midrule
                    
            %        L2R \citep{FSR} &  66.51 & 78.93 & 85.19  & 33.38 & 44.44 & 53.73  \\
            %        MWN \citep{FSR} &  68.91 & 80.06 & 87.84  & 37.91 & 46.74 & 58.46  \\
                    %FaMUS-CE\citep{Famus}  &  - & 83.15 & 89.39 &  - &  49.56 & 60.42  \\
                    %FaMUS-LDAM\citep{Famus} &  - & 83.32 & 87.90 &  - &  49.93 & 59.03  \\
            %        FSR-DF \citep{FSR} &      66.15 & 79.78 & 88.15 &  36.74 & 44.43 & 55.60  \\
            %        FSR \citep{FSR} &  67.76 & 79.17 & 87.40 &  35.44 & 42.57 & 55.45  \\
            %        \rowcolor{gray!30} \textbf{Ours} & \textbf{74.91}   &  \textbf{84.43} & \textbf{90.81}  &  \textbf{41.52} & \textbf{51.35} & \textbf{62.07} \\
            %        \bottomrule
            %    \end{tabular}
            %\end{table}
        %\vspace{-1em}
        
        \subsection{Imbalanced noisy-label learning}
            We evaluate our proposed method in the setting that combines class imbalance and label noise from~\citep{FSR}. We follow the same experimental configuration by adding 20\% and 40\% symmetric noise to the CIFAR10 imbalanced datasets with different imbalance ratios (10, 50 and 200).
            The results in \cref{tab:table_7} show that INOLML outperforms other approaches by a large margin. 
            This result is even more remarkable when studying the  noise rate of 40\%. 
            For CIFAR100, we do not show results with imbalance ratio $>10$ since for larger imbalance ratios, it was impossible to build validation sets with 10 samples per class for the minority classes.
            %they did not allow us to build validation sets with 10 samples per class, and the validation set will guarantee to contain noise regardless of how we select the samples. \gustavo{COMMENT:We need to discuss this limitation!}
            Nevertheless, for the two CIFAR100 problems, our method shows substantially better results than previous SOTA methods.
            Our method can, therefore, be considered the new SOTA for this imbalanced noisy-label learning benchmark with Resnet32.
            %We  provide details on the hyper-parameters used, e.g. \(k\), in \cref{sec:hyperparameter_for_imbalance}.

            \begin{table}[t]
                \centering
                \caption{Test accuracy (\%) of  INOLML and other SOTA methods on CIFAR10 and CIFAR100 imbalanced learning mixed with symmetric noise. The reported results are  from~\citep{FSR} and \citep{ROLT}.
                % All results are the average of 3 runs.
                }
                \label{tab:table_7}
                % \scalebox{0.8}{
                \begin{tabular}{l l l l l l l l l }
                    \toprule
                    \bfseries Dataset & \multicolumn{6}{c}{\bfseries CIFAR10} &\multicolumn{2}{c}{\bfseries CIFAR100} \\
                    \cmidrule(lr){2-7} \cmidrule(lr){8-9}
                    \bfseries Noise ratio& \multicolumn{3}{c}{\bfseries 0.2} &  \multicolumn{3}{c}{\bfseries 0.4} & \bfseries 0.2 & \bfseries 0.4\\
                    \cmidrule(lr){2-4} \cmidrule(lr){5-7} %\cmidrule(lr){8-9}
                    \bfseries Imbalance ratio & \bfseries 10 & \bfseries 50 & \bfseries 200 & \bfseries 10 & \bfseries 50 & \bfseries 200 &\bfseries 10 & \bfseries 10  \\
                    % \cmidrule(lr){2-4} \cmidrule(lr){5-7} \cmidrule(lr){8-9}
                    \midrule
                    CRUST  &  65.7 & 41.5 & 34.3 &  59.5 & 32.4 & 28.8 & - &- \\
                    LDAM  &  82.4   & -  & -  &  71.4 & -  &  - &48.1   &36.7 \\
                    LDAM-DRW  &  83.7   & -  & -  &  74.9 & -  & -  &50.4   & 39.4 \\
                    BBN  &  80.4 & -  & -  &  70.0 & -  & -  & 47.9 &35.2 \\
                    HAR-DRW  &  82.4 & -  & -  &  77.4 & -  & -  & 46.2 & 37.4 \\
                    ROLT-DRW  &  85.5   & -  & -  &  82.0 & -  &  - &52.4   &46.3 \\
                    FSR  &  85.7 & 77.4 & 65.5 &  81.6 & 69.8 & 49.5 & - & -   \\
                    \textbf{INOLML} & \(\mathbf{90.1}\) & \(\mathbf{80.1}\) & \(\mathbf{66.6}\) &  \(\mathbf{89.1}\) & \(\mathbf{78.1}\) & \(\mathbf{61.6}\) & \(\mathbf{59.8}\)& \(\mathbf{56.1}\) \\
                    \bottomrule
                \end{tabular}
                % }
            \end{table}

        \subsection{Open-set noise}
        
            Open-set noise occurs when training images may belong to classes outside the \(C\) classes in \(\mathcal{D}\). We consider 3 benchmarks using CIFAR10 contaminated with images from CIFAR100 and ImageNet~\citep{lee2019robust}.
            \cref{tab:openset_noise} shows results from INOLML, Distill~\citep{Distill_noise} and other meta-learning methods~\citep{assmymetric,L2W,FSR}, where INOLML achieves the best performance in all cases. Comparing to Distill, INOLML is 0.5\% to 1\% better, despite the selected validation set \(\mathcal{D}^{(v)}\) being largely contaminated with noisy samples (up to 40\%), as shown in Figure~\ref{fig:probe_acc_open}.
            However, such performance on open noise contrasts with our observation in the 80\% symmetric noise settings, where just 30\% noise rate in \(\mathcal{D}^{(v)}\) degrades the performance of INOLML, compared to Distill model (\cref{tab:table_2}). 
            Such difference
            might be attributed to the out-of-distribution characteristic of open-set noise. As open-set noisy-label datasets contain samples that do not belong to the set of known classes, such samples might help regularise the learning on mislabelled data, reducing the effect of confirmation bias, resulting in a better performance. %As the results, our methods is robust to open-set  noise scenario.
            
            \begin{table}[t]
                \centering
                \caption{Test accuracy (\%) of  INOLML and previous methods in open-set noise using WideResnet28-10 with 10 samples per class for validation.
                }
                \label{tab:openset_noise}
                % \scalebox{0.8}{
                \begin{tabular}{l l l l}
                    \toprule
                    \bfseries Method & \bfseries ImageNet  & \bfseries CIFAR100  & \bfseries BOTH  \\
                    \midrule
                    RoG~\citep{assmymetric} &  83.4 & 87.1 & 84.4  \\
                    L2R~\citep{L2W} &  81.8 & 81.8 & 85.0  \\
                    Distill~\citep{Distill_noise} &  94.0 & 92.3 & 93.0  \\
                    \textbf{INOLML} &  \(\mathbf{94.5} \pm \mathbf{0.1}\) & \(\mathbf{93.6} \pm \mathbf{0.0}\) & \(\mathbf{93.6} \pm \mathbf{0.1}\) \\
                    \bottomrule
                \end{tabular}
                % }
                %\vspace{-1em}
            \end{table}

            \subsection{Real-world datasets}

            \cref{tab:sota_results_CNWL,tab:sota_results_web} show the  results of INOLML and other SOTA approaches on real-world datasets.  ~\cref{tab:sota_results_web} shows the performance on mini-WebVision with 2 popular models: Resnet50 (RN50) and InceptionResnetV2 (InRN), while \cref{tab:sota_results_CNWL}  shows results on four different noise ratios evaluated on Red mini-ImageNet with 32 \(\times\) 32 images with 2 popular models: Resnet18 (RN18) and InceptionResnetV2 (InRN). INOLML is the new SOTA on mini-WebVision with Resnet50 model %and is competitive with the best method~\citep{ortego2021multi} on 
            and Red mini-ImageNet. We note that our  method is more efficient in terms of memory footprint than most of Co-training based approaches~\citep{li2020dividemix,propmix,Famus} evaluated on Red Mini-ImageNet since we use only a single PreAct Resnet18 with meta-learning instead of two separate PreAct Resnet18. On mini-WebVision benchmark, our method can also achieve significantly better performance despite using a smaller architecture in terms of the number of parameters compared to previous methods.

            \begin{table}[t]
                \centering
              \caption{Prediction accuracy (\%) on the real-world dataset mini-WebVision with Resnet50, evaluated on Webvision and ImageNet test sets; the results of other methods are  from \citep{FSR,propmix} or from their original papers.}
                \label{tab:sota_results_web}
                \scalebox{1.0}{
                  \begin{tabular}{l c c  c c}
                        \toprule
                        \multirow{2}{*}{\bfseries Method} &   \multirow{2}{*}{\bfseries Base} & \multirow{2}{*}{\bfseries \shortstack{ Num. \\ Params} }  &\multirow{2}{*}{\bfseries WebVision} & \multirow{2}{*}{\bfseries ImageNet} \\
                        %\cmidrule(lr){4-5} \cmidrule(lr){6-7}
                        %&  & & \bfseries top-1 & \bfseries top-5 & \bfseries top-1 & \bfseries top-5  \\
                        & & & & \\
                        \midrule
                        HAR & InRN  & 56M  &  75.5 &   57.4    \\ %\citep{Imbalance_noisy}
                        %D2L   &  62.7& 84.0  & 57.8 & 81.4  \\ %\citep{ma2018dimensionality}
                        %Co-teaching & & &    &  63.6 & 85.2 &  61.5 & 84.7   \\ %\citep{han2018co}
                        %Iterative-CV & & &    &  65.2 & 85.3 &  61.6 & 85.0   \\ %\citep{Chen2019UnderstandingAU}
                        %MentorNet & & &   &  63.0 & 81.4 &  63.8 & 85.8  \\ %\citep{MentorNet}
                        %CRUST &  &  & 72.4 & 89.6 &67.4 & 87.8 \\
                        %CRUST & \multirow{2}{*}{72.4} & \multirow{2}{*}{89.6} & \multirow{2}{*}{67.4} & \multirow{2}{*}{87.8} \\
                        %\citep{CRUST} & & & & \\
                        GJS & RN50  &  23M   &  78.0    & 74.4  \\ %\citep{GJS}
                        MW-Net & InRN  & 56M   &  74.5  & 72.6 \\ % \citep{MetaWeightNetLA}
                UNICON & 2$\times$InRN  & 112M  &  77.6   & 75.3 \\
                MOIT  & RN18 & 11M &  78.8  & - \\
        %MOIT & &   &  78.8  & -  & - & - \\
                        
                         SSR &  InRN  & 56M & 80.9 & 75.8  \\
                        C2D &  2$\times$ RN50 & 46M  & 79.4  &  \( \mathbf{78.6}\) \\
                        Bayesian DivideMix++   & 2$ \times$ RN50 & 23M   &  80.12    & 78.51  \\
                         NCR   & RN50 & 23M & 80.5 & -  \\
                         BtR   & InRN & 56M & 80.9 & 76.0  \\
                         CC   & InRN & 56M & 79.4 & 76.1  \\
                         TLC  & RN50 & 23M   &  79.1   & 75.4  \\ 
                        FSR  & RN50 & 23M   &  74.9    & 72.3  \\ %\citep{FSR}
                        \textbf{INOLML} & RN50 & 23M   & \(\mathbf{81.7}\)  & 78.1 \\
                        \bottomrule
                    \end{tabular}
                    }
            \end{table}

            \begin{table}[t]
                \centering
              \caption{Prediction accuracy (\%) on the real-world dataset Red mini-ImageNet dataset. The results of other methods are  from \citep{FSR,propmix} or from their original papers.}
                \label{tab:sota_results_CNWL}

                \scalebox{1.0}{
                  \begin{tabular}{l l c c c c c}
                        \toprule
                        \multirow{2}{*}{\bfseries Method} &   \multirow{2}{*}{\bfseries Backbone} & \multirow{2}{*}{\bfseries \shortstack{ \textnumero~parameters} } & \multicolumn{4}{c}{\bfseries Noise ratio} \\
                        \cmidrule{4-7} 
                        & & (millions) & \bfseries 0.2 & \bfseries 0.4 & \bfseries 0.6 & \bfseries 0.8 \\
                        \cmidrule{1-7}
                         CE  & 2$\times$ RN18 & 22 & 47.36 & 42.70 & 37.30 & 29.76  \\ %Cross entropy\citep{propmix}
                        Mix Up  & 2$\times$ RN18 & 22  & 49.10 &  46.40 & 40.58 & 33.58  \\ %\citep{propmix}
                        DivideMix  & 2$\times$ RN18 & 22  & 50.96 & 46.72 & 43.14 & 34.50  \\ %\citep{li2020dividemix}
                        MentorMix  & InRN & 56 & 51.02 & 47.14 & 43.80 & 33.46  \\ %\citep{jiang2020beyond}
                        PropMix  & 2$\times$ RN18 & 22 & 61.24 & 56.22 & 52.84 & 43.42 \\ %\citep{propmix}
                        %MOIT  & 63.14 & \(\mathbf{60.78}\) & - & \(\mathbf{45.88}\) \\ 
                        %\citep{ortego2021multi}
                        FaMUS  & 2$\times$ RN18 & 22 & 51.42 & 48.06 & 45.10 & 35.50  \\ %\citep{Famus}
                        InstanceGM & 2$\times$ RN18 & 22 & 58.38 & 52.24 & 47.96 & 39.62  \\ %\citep{Famus}
                        InstanceGM-SS & 2$\times$ RN18 & 22 & 60.89 & 48.06 & 45.10 & 35.50  \\ %\citep{Famus}

                        \textbf{INOLML}  & RN18 & 11 & \(\mathbf{63.23}\) & \(\mathbf{58.21}\) & \(\mathbf{53.39}\) & \textbf{45.32}  \\
                        %\rowcolor{gray!30} \textbf{Ours} & \textbf{62.03} & \textbf{56.91} & 51.29 & \textbf{43.92}  \\
                        \bottomrule
                    \end{tabular}
                    }
                    %\end{center}
                %\vspace{-1em}
            \end{table}
    \section{Ablation studies}
%\vspace{-0.5em}
    %\subsection{How does the algorithm work without diversity factor/similarity constraint/improving the pseudo clean subset?}

        This section studies the factors affecting the optimization in~\eqref{eq:validation_set_criteria}.
        In the lower-lever optimization of~\eqref{eq:validation_set_criteria}, the function $\mathsf{Info(.)}$ selects samples that maximise the training sample weight \(\omega\) from~\eqref{eq:gradient_w}, as well as samples that maximise the maximum \say{information content}.
        % that each training sample can get from any  sample in the clean validation set.  
        We, therefore, carry out an ablation study about the importance of this factor by replacing $\mathsf{Info(.)}$ in~\eqref{eq:validation_set_criteria} with the $\mathsf{MaxWeight(.)}$ in~\cref{eq:naive_optimization} and show the results in \cref{tab:ablatant}. 
        We also study the role of
        the $\mathsf{Clean}(.)$ utility function
        in~\eqref{eq:validation_set_criteria} by optimising only the lower-level of~\eqref{eq:validation_set_criteria} (see $\mathsf{Info}(.)$ \textbf{only}). This ablation study is conducted on CIFAR10 and CIFAR100 under 40\% asymmetric noise and 20\% symmetric noise with imbalanced data.
        Overall, each component improves model performance compared to naively optimising the average weight in~\eqref{eq:gradient_w}.  
        %Adding $\mathsf{Info}(.)$ and  $\mathsf{Clean}(.)$ show substantial improvements.
        Naively selecting samples based on~\eqref{eq:gradient_w}  facilitates the overfiting of the noisy-label samples, leading to confirmation bias. 
        The framework mitigates this problem by using $\mathsf{Clean}(.)$ limiting the noise in the clean validation set, while $\mathsf{Info}(.)$ prevents the gradient to go toward a single wrong direction.

        As shown in \cref{tab:ablatant}, the impact of $\mathsf{Clean}(.)$ varies across different settings. Consequently, applying a uniform optimization framework for all noise types, noise rate and imbalanced ratio may result in unnecessary overhead and potentially suboptimal performance. A promising future direction could involve designing an adaptive framework that automatically determines the optimal training strategy by estimating the class imbalance and noise rate, thereby minimizing framework overhead and optimizing performance.

\begin{figure}
        % \vspace{-0.5em}
        \centering
         \includegraphics[width=1\textwidth]
         {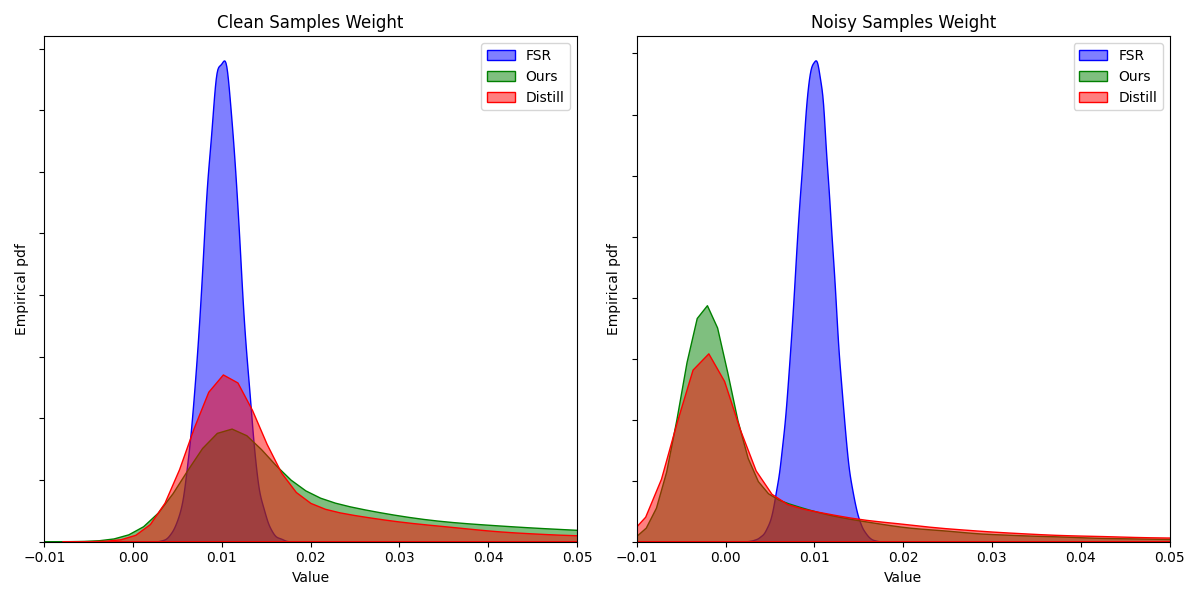} 
        \vspace{-0.5cm}
        \caption{Weight distribution of samples from different data reweighting methods, under the setting CIFAR100 with 0.8 uniform noise ratio. 
        %The color bars on the right-hand side denote the differences between pixel values of the warm-up and generated images.
        }
    \label{fig:weight}
    \vspace{-5mm}
    \end{figure}

        \begin{table}[t]
            \centering
            \caption{Test accuracy (\%) on CIFAR10 and CIFAR100 under asymmetric and imbalanced noisy-label problems, where IR denotes the imbalance ratio. The $1^{st}$ row shows the results of the optimization of the average of weight (col. \textbf{Average Weight in~\eqref{eq:gradient_w}})   instead of \eqref{eq:validation_set_criteria}. The  $2^{nd}$ row shows the results of optimising the lower part of \eqref{eq:validation_set_criteria} (col. \textbf{$\mathsf{Info}(.)$ Only}) without the upper part of \eqref{eq:validation_set_criteria}  $\mathsf{Clean}(.)$. The last row (\textbf{Whole \eqref{eq:validation_set_criteria}}) shows our final model result.}
            \label{tab:ablatant}
            %\begin{tabular}{l@{\hskip 1em} c@{\hskip 1em} c@{\hskip 1em} c@{\hskip 1em} c@{\hskip 1em} c@{\hskip 1em} c@{\hskip 1em} c@{\hskip 1em} c@{\hskip 1em} c@{\hskip 1em} c}
                %\toprule
                %\multirow{3}{*}{\textnumero} & \multirow{3}{*}{\bfseries Weight} & \multirow{3}{*}{\bfseries Div.} & \multirow{3}{*}{\bfseries Sim.} &  \multicolumn{4}{c}{\bfseries Sym.} & \multicolumn{1}{c}{\bfseries Asym.} & \multicolumn{2}{c}{\bfseries Semantic}  \\
                %\cmidrule{5-11} 
                %&  & &  & \multicolumn{2}{c}{CIFAR10} & \multicolumn{2}{c}{CIFAR100} & \multicolumn{1}{c}{CIFAR10} & \multicolumn{1}{c}{CIFAR10} & \multicolumn{1}{c}{CIFAR100}  \\
                %\cmidrule{5-11} 
                    %& & & &  0.4 & 0.8 & 0.4 & 0.8 & 0.4 & 0.34 & 0.37 \\
                %\midrule
                %Random selection & 95.5 & 91.8 & 79.6 & 73.6 & 94.5  \\ \hline
                %Random Selection  & - &  - & 80.1 & 73.6 & 94.1  \\ \hline
                %Max Weight
                %1 & \checkmark &  & & 95.9 & 93.8  & 81.0 & 74.0 & 91.0 & 85.8 & 72.4  \\ 
                
                %Max Weight + Pseudo label 
                %2 & \checkmark & \checkmark & & & 96.0 &  94.0 & 81.2 & 74.3 & 90.5 & 86.5 &72.5  \\ 

                %Max Weight + Pseudo label + Diversity  
                %2 & \checkmark  & \checkmark & & 96.0 &  94.6 & 81.1 &  74.2 & 92.1  & 86.5  & 72.5 \\
                
                %Max Weight + Diversity +  Similarity constraint
                %4 & \checkmark & \checkmark & \checkmark  & 96.0 &  94.5 & 81.3 & 74.3 & 94.0  & 86.6  &72.5 \\ 

                %Max Weight + Pseudo label + Diversity + Similarity constraint 
                %3 & \checkmark  & \checkmark & \checkmark & 96.2 &  94.6 & 81.3 & 74.5 & 94.1 & 87.0 & 73.0  \\
                %\bottomrule
            %\end{tabular}
            \scalebox{0.9}{
            \begin{tabular}{ c c c |c c |c c c|}
                \toprule
                 \multirow{5}{*}{\bfseries \shortstack[l]{ \shortstack[c]{Replace $\mathsf{Info}$ \\ with  $\mathsf{MaxWeight}$}}} & \multirow{5}{*}{\bfseries \shortstack[c]{$\mathsf{Info}(.)$ \\ Only }} & \multirow{5}{*}{\bfseries \shortstack[c]{Equation \\  \eqref{eq:validation_set_criteria} }} &  \multicolumn{2}{|c|}{\bfseries Asymmetric} & \multicolumn{3}{c|}{\bfseries  Symmetric}  \\
                \cmidrule{4-8} 
                & &  &  \multicolumn{2}{|c|}{CIFAR10} & \multicolumn{3}{c|}{CIFAR10}  \\
                \cmidrule{4-8} 
                & & & WRN & RN29 & RN32 & RN32 & RN32  \\
                %\cmidrule{1-8} 
                \multicolumn{3}{c|}{
                %\bfseries Imb. Ratio $\rightarrow$
                } 
                & IR=1 & IR=1 & IR=10 & IR=50 & IR=200 \\
                    %& & & &  0.4 & 0.8 & 0.4 & 0.8 & 0.4 & 0.34 & 0.37 \\
                \midrule
                %Random selection & 95.5 & 91.8 & 79.6 & 73.6 & 94.5  \\ \hline
                %Random Selection  & - &  - & 80.1 & 73.6 & 94.1  \\ \hline
                %Max Weight
                \checkmark &  & & 91.0 &56.6 & 68.8 & 37.6 & 23.4   \\ 
                
                %Max Weight + Pseudo label 
                %2 & \checkmark & \checkmark & & & 96.0 &  94.0 & 81.2 & 74.3 & 90.5 & 86.5 &72.5  \\ 

                %Max Weight + Pseudo label + Diversity  
                 & \checkmark & & 92.1 & 89.3  & 89.0 & 79.1  & 65.9  \\
                
                %Max Weight + Diversity +  Similarity constraint
                %4 & \checkmark & \checkmark & \checkmark  & 96.0 &  94.5 & 81.3 & 74.3 & 94.0  & 86.6  &72.5 \\ 

                %Max Weight + Pseudo label + Diversity + Similarity constraint 
                  &  & \checkmark & 94.1 &89.8 & 90.1 & 80.1 & 66.6  \\
                \bottomrule
            \end{tabular}
            }
        % \vspace{-1em}
        \end{table}
        To demonstrate the effectiveness of INOLML, we measure the distributions of the sample weights $\omega$ during the training process compared to other meta reweighting methods, as demonstrated in Figure~\ref{fig:weight}.
        Recall that INOLML targets the maximisation of $\omega$ for training samples of high utility, so it is important to measure how $\omega$ progresses during training. The evaluation is conducted on CIFAR-100 with 80\% symmetric noise setting, and we compare against FSR~\citep{FSR} and Distill~\citep{Distill_noise}, as they are the most closely related frameworks to our approach. The results in Figure~\ref{fig:weight} show that overall, our method provides higher weight for clean samples and lower weight for noisy samples, compared to other meta reweighting methods such as FSR~\citep{FSR} or Distill~\citep{Distill_noise}, demonstrating the effectiveness of our method.
        
        We also compare the validation set produced  from~\eqref{eq:validation_set_criteria} with sets built with random sampling and most confident sampling based on the highest confidence scores. %\gustavo{how is the most confident sampling done?} \dung{I add a bit detail there} on Resnet29.  
        %We also show the effectiveness of our methods compared to the random selection of clean sample and sampling the most confident samples by comparing the models's performance using these 3 different selection methods using Resnet29 model in Ficgure.
         Figure~\ref{fig:selection_mostconfident} shows that the most confident sampling has inferior results compared to random sampling. Nevertheless, INOLML with built-in high utility validation set formation shows the best results.

        %selecting the most confident samples can lead to biased that degrade the performance of the model compared to a randomly selected clean validation set despite it avoiding majority of noisy samples, while our methods show superior result with a noisy validation set. 
        Traditional meta-learning approaches \citep{L2LWS,MetaWeightNetLA} always keep the clean validation set separate from the training set, while our method iteratively extracts $D^{(v)}$ from the training set.
        It can be argued that this non-separation of the training and validation sets can cause confirmation bias to happen during training. 
        Hence, we evaluate our approach in a scenario where the candidate samples to form the validation set is always separate from the training set during training.
        However, our results show
        that such separate validation set causes a $2\%$ drop in accuracy, on average.
        This can be explained by the smaller size of the training set and the restriction in potential choices for validation samples.
        %e reason is due to the training set get smaller, while the potential choices for the clean validation set get limited as well.

        % \begin{figure}[t]
        %     \centering
        %     \includegraphics[width = 0.99\linewidth]{img/ablatant_study_8.pdf}
        %     \caption{Accuracy (\%) of our INOLML using different sample selection methods.
        %     } \label{fig:selection_mostconfident}
        % \end{figure}

     \begin{figure}
            % \vspace{-1em}
            \centering
            \begin{tikzpicture}
                \begin{axis}[
                    height=0.25\linewidth,
                    width=0.5\linewidth,
                    ybar=1pt,
                    ylabel={Accuracy},
                    ylabel style={font=\footnotesize, yshift=-0.5em},
                    yticklabel style={font=\footnotesize},
                    ymax=100,
                    bar width=15pt,
                    symbolic x coords={{CIFAR10 0.4}, {CIFAR10 0.8}, {CIFAR100 0.8}},
                    xtick=data,
                    xticklabel style = {font=\footnotesize},
                    nodes near coords,
                    nodes near coords align={vertical},
                    nodes near coords style={font=\scriptsize, /pgf/number format/.cd, fixed zerofill, precision=1},
                    enlarge x limits=0.375,
                    scale only axis,
                    legend style={draw=none, font=\scriptsize, yshift={0.125em}},
                    legend pos=north east,
                    legend image post style={scale=1},
                    legend cell align={left}
                ]
                    \addplot[style={fill=MidnightBlue, draw=none}, error bars/.cd, y dir=both, y explicit] coordinates {
                        (CIFAR10 0.4, 88.6)
                        (CIFAR10 0.8, 79.1)
                        (CIFAR100 0.8, 50.7)
                    };
                    \addplot[style={fill=BurntOrange, draw=none}, error bars/.cd, y dir=both, y explicit] coordinates {
                        (CIFAR10 0.4, 89.2)
                        (CIFAR10 0.8, 87.0)
                        (CIFAR100 0.8, 57.5)
                    };
                    \addplot[style={fill=PineGreen, draw=none}, error bars/.cd, y dir=both, y explicit] coordinates {
                        (CIFAR10 0.4, 91.0)
                        (CIFAR10 0.8, 87.9)
                        (CIFAR100 0.8, 59.2)
                    };
                    \legend{{Most confidence}, {Random clean}, INOLML};
                \end{axis}
            \end{tikzpicture}
            % \vspace{-0.5em}
            \caption{Accuracy (\%) of our INOLML using different sample selection methods under uniform label noises.}
            \label{fig:selection_mostconfident}
            %\vspace{-1em}
            
        \end{figure}
        
        %Recall that INOLML targets the maximisation of $\omega$ for training samples of high utility, so it is important to measure how $\omega$ progresses during training. 
        %This is evaluated on CIFAR-10 with 80\% symmetric noise. The results in \cref{fig:weight} show that the mean value of $\omega$ (labelled as mean weight) increases for the clean samples (left graph) and decreases for noisy samples (right graph).
        %Also, the selection of validation samples using our criteria (blue curve) produces higher values of $\omega$ for the clean samples than randomly selecting validation samples (orange curve). For noisy samples, our validation set selection leads to smaller values of $\omega$ than the random validation selection methods \gustavo{Which figure is this referring to?}. %\dung{I fixed the chart.}

        % \subsection{What is the extra time required for using the algorithm together with the model?}
        %     \vspace{-0.5em}
            
            A final discussion point is the time needed for the INOLML training. Using CIFAR10 with uniform noise, 
            %\gustavo{which noisy-label problem?} \dung{I believe it does not matter, the noise rate has nothing to do with the training time unless we tune the hyperparameter for each noise rate (I use the same hyperparameter though)}, 
            the Distill model takes around 5 and 29 hours to train the Resnet29 and WideResnet28-10 models, respectively. When integrating our method with Distill, training  slightly increases to 5.5 hours on Resnet29 and 31 hours on WideResnet28-10. Hence, our algorithm adds around 10\% to the training time for optimising the validation set, which happens once per epoch.
            The experiment above was conducted on a single NVIDIA V100 GPU.
            We also compare the training time of our approach with other recently proposed methods using the PreAct Resnet18 model on CIFAR100 dataset with 40\% uniform noise using a single V100 GPU.  While our approach takes $10.5$h, DivideMix~\citep{li2020dividemix} takes 8.25 hours, CausalNL\citep{causalnl} takes around 12.5 hours, and MOIT~\citep{ortego2021multi} takes 8 hours, indicating that our approach has competitive training time compared to recently proposed  methods.

            % \begin{figure}[t]
    %         \centering
    %         \includegraphics[width=0.4\textwidth,height=0.15\textwidth]{content_WACV_revise/fixed_ablatant_plot.png} 
    %         \vspace{-0.75em}
    %         \caption{Mean $\omega$ given the training iterations, for the selected clean (left side) and noisy (right side) samples using our validation set selection (blue curves) compared to a random validation selection ( orange curves) on CIFAR-10 with 0.8 symmetric noise.}
    %         \label{fig:weight}
    %         \vspace{-1.25em}
    % \end{figure}

    \section{Conclusion}
\label{sec:conclusion}
\noindent
 We presented a novel meta-learning approach, called INOLML, that automatically and progressively selects a pseudo-clean validation set from a noisily-labelled training set. 
 This selection is based on our proposed validation set utility criteria that take into account sample informativeness, class-balanced distribution, and label cleanliness. Our proposed method is more effective and practical than prior  meta-learning approaches since we do not require manually-labelled samples to be included in the validation set. 
 Compared with other meta-learning approaches that do not require a manually labelled validation set (e.g. FSR or FaMUS), INOLML has demonstrated to be more robust to high noise rate problems and able to achieve SOTA results on several synthetic and real-world label noise benchmarks.
 In fact, INOLML has SOTA results in mini-WebVision, Red mini-ImageNet, open-set noise, long-tailed + symmetric noise for CIFAR-10/-100, symmetric and asymmetric CIFAR-10/-100, and imbalanced learning, with substantial improvements (e.g., CIFAR-10 80\% symmetric, CIFAR-100 40\% symmetric, imbalanced and noisy-label imbalanced benchmarks, and mini-WebVision).

 A limitation of our approach is that the model can suffer from confirmation bias as it is based on a single network. 
 As future work, we will tackle this problem by incorporating co-teaching in our meta-learning algorithm. 
 %Another potential improvement is the random selection of $N$ samples per class in \cref{sec:training_procedure}, which may not contain representative samples for the datasets, and can be replaced by any other pivoting selection methods.
 Another limitation is the greedy and complex bi-level optimization to form the validation set in~\eqref{eq:validation_set_criteria}, which can be improved in two ways: 1) the complexity can be reduced by replacing the bi-level optimization with a single-level optimization, and 2) the greedy strategy can be replaced by a relaxation method to solve the combinatorial optimization problem.
 Additionally, optimising the clean validation set once per epoch is not ideal since the validation set can be outdated by the end of epoch. This issue will be addressed by updating the clean validation set more regularly.
 Finally, another point missing from this paper is a theoretical analysis of the proposed criteria to characterise the utility of the meta-learning validation set. 
 In particular, we plan to study why the relaxation of the assumption of clean and balanced validation set made by~\citet{L2W} in~\eqref{eq:gradient_w} still works for our validation set that has pseudo-clean, balanced and informative samples.
 \subsubsection*{Acknowledgments}
 G.C. and C.N. are supported by the Engineering and Physical Sciences Research Council (EPSRC) through grant EP/Y018036/1.

\bibliography{main}
\bibliographystyle{tmlr}

\newpage
\appendix
\section{Real world noise results}
\begin{table}[ht]
                \centering
              \caption{Prediction accuracy (\%) on real-world datasets. \emph{(left):} WebVision with Resnet50, evaluated on Webvision and ImageNet test sets; and \emph{(right):} Red Mini-ImageNet. The results of other methods are  from \citep{FSR,propmix} or from original papers.}
                \label{tab:sota_results}
            \begin{minipage}[t]{0.48\linewidth}\centering

                \scalebox{1.0}{
                  \begin{tabular}{l c  c c c}
                        \toprule
                        \multirow{2}{*}{\bfseries Method} &   \multicolumn{2}{c}{\bfseries WebVision} & \multicolumn{2}{c}{\bfseries ImageNet} \\
                        \cmidrule(lr){2-3} \cmidrule(lr){4-5}
                        & \bfseries top-1 & \bfseries top-5 & \bfseries top-1 & \bfseries top-5  \\
                        \midrule
                        HAR  &  75.5 & 90.7 &  57.4 & 82.4   \\ %\citep{Imbalance_noisy}
                        %D2L   &  62.7& 84.0  & 57.8 & 81.4  \\ %\citep{ma2018dimensionality}
                        Co-teaching   &  63.6 & 85.2 &  61.5 & 84.7   \\ %\citep{han2018co}
                        Iterative-CV    &  65.2 & 85.3 &  61.6 & 85.0   \\ %\citep{Chen2019UnderstandingAU}
                        MentorNet  &  63.0 & 81.4 &  63.8 & 85.8  \\ %\citep{MentorNet}
                        CRUST & 72.4 & 89.6 &67.4 & 87.8 \\
                        %CRUST & \multirow{2}{*}{72.4} & \multirow{2}{*}{89.6} & \multirow{2}{*}{67.4} & \multirow{2}{*}{87.8} \\
                        %\citep{CRUST} & & & & \\
                        GJS   &  78.0  & 90.6  & 74.4 & 91.2 \\ %\citep{GJS}
                        MW-Net  &  74.5 & 88.9  & 72.6 & 88.8 \\ % \citep{MetaWeightNetLA}
                UNICON  &  77.6  & 93.4  & 75.3 & \(\mathbf{93.7}\) \\
        MOIT  &  78.8  & -  & - & - \\
                        FSR  &  74.9  & 88.2  & 72.3 & 87.2 \\ %\citep{FSR}
                        
                        \textbf{INOLML}  & \(\mathbf{81.7}\) & \(\mathbf{93.8}\)  & \(\mathbf{78.1}\) & \(92.9\) \\
                        \bottomrule
                    \end{tabular}
                    }
                    %\end{center}
                    \end{minipage}
            \begin{minipage}[t]{0.48\linewidth}\centering

                \scalebox{1.0}{
                  \begin{tabular}{l c c c c}
                        \toprule
                        \multirow{2}{*}{\bfseries Method} & \multicolumn{4}{c}{\bfseries Noise ratio} \\
                        \cmidrule{2-5} 
                        & \bfseries 0.2 & \bfseries 0.4 & \bfseries 0.6 & \bfseries 0.8 \\
                        \cmidrule{1-5}
                         CE & 47.36 & 42.70 & 37.30 & 29.76  \\ %Cross entropy\citep{propmix}
                        Mix Up  & 49.10 &  46.40 & 40.58 & 33.58  \\ %\citep{propmix}
                        DivideMix  & 50.96 & 46.72 & 43.14 & 34.50  \\ %\citep{li2020dividemix}
                        MentorMix & 51.02 & 47.14 & 43.80 & 33.46  \\ %\citep{jiang2020beyond}
                        PropMix & 61.24 & 56.22 & 52.84 & 43.42 \\ %\citep{propmix}
                        %MOIT  & 63.14 & \(\mathbf{60.78}\) & - & \(\mathbf{45.88}\) \\ %\citep{ortego2021multi}
                        FaMUS & 51.42 & 48.06 & 45.10 & 35.50  \\ %\citep{Famus}
                        \textbf{INOLML} & \(\mathbf{63.23}\) & \(\mathbf{58.21}\) & \(\mathbf{53.39}\) & \textbf{45.32}  \\
                        %\rowcolor{gray!30} \textbf{Ours} & \textbf{62.03} & \textbf{56.91} & 51.29 & \textbf{43.92}  \\
                        \bottomrule
                    \end{tabular}
                    }
                    %\end{center}
                    \end{minipage}
                %\vspace{-1em}
            \end{table}
\section{Implementation Details}
\label{sec:implementation_details_appendix}
All CIFAR experiments use batches of size 100, which are trained on a single GPU. Similar to the Distill noise model~\citep{Distill_noise}, we use \(p=5, k=20\) for CIFAR experiments, except the ones with the imbalance setting.

For Red Mini-ImageNet experiments, we trained the model on a single  GPU with batches of size 100, with \(p=5, k=10\).

For the WebVision experiment, we use \(p=4, k=8\) with 4 NVIDIA V100 GPU and batches of size 16.  All experiments use \(N=200, K=50, \kappa = 0.9\).

In practice, to reduce the computational cost of the optimisation in~\eqref{eq:validation_set_criteria}, we replace the pseudo-clean set $\mathcal{D}^{(c)}$ with the following
    %depends on the actual (hidden) proportion of clean samples in the pseudo clean set $\mathcal{D}^{(c)}$, while the efficiency depends on the size of $\mathcal{D}^{(c)}$. Hence, to reduce computational cost, the selection of the validation set in~\eqref{eq:validation_set_criteria} uses the
    subset:
    \begin{equation*}
    \begin{aligned}[b]
        \widetilde{\mathcal{D}}^{(c)} & = \left\{ \vphantom{\max_{k\in\{1,...,C\}}} (\mathbf{x}_{i}, \mathbf{y}_{i}): (\mathbf{x}_{i}, \mathbf{y}_{i}) \in \mathcal{D}^{(c)}  \wedge \, \operatorname*{argmax}_{k\in\{1,...,C\}} \mathbf{y}_{i}(k) = \operatorname*{argmax}_{k\in\{1,...,C\}} \tilde{\mathbf{y}}_{i}(k) \right\}.
    \end{aligned}
    \end{equation*}
    In summary, we aim to progressively refine the pseudo clean set $\mathcal{D}^{(c)}$, making it cleaner over time.
    $\widetilde{\mathcal{D}}^{(c)}$ is a subset of $\mathcal{D}^{(c)}$, containing $N$ randomly-selected samples \((\mathbf{x}_{i}, \mathbf{y}_{i})\) of each class in $\mathcal{D}^{(c)}$ that have their observed labels $\mathbf{y}_{i}$ consistent with the corresponding moving average robust label computed with the average prediction over the last $E$ epochs, as in:
    \begin{equation*}
        \tilde{\mathbf{y}}_i = \kappa \tilde{\mathbf{y}}_i +  (1-\kappa)\frac{1}{E} \sum_{e=1}^{E} f_{\theta}(\mathbf{x}_i),
    \end{equation*}
    with \(\kappa \in [0, 1]\) being a hyper-parameter. 
\section{Additional Results of Symmetric Noise on CIFAR Datasets}
\label{sec:additional_symmetric_noise_results}
        We provide additional symmetric noise results of our proposed method and the Distill model~\citep{Distill_noise} in Table \ref{tab:table_1}. Note that our method is markedly better than Distill, particularly for the simpler model (RN29) with few samples per class (1 and 5) in the validation set. For the more complex model (WRN) and large validation set (10 samples per class), our method is still better than Distill, except for CIFAR100 at 0.8 symmetric noise rate. 

       As shown in ~\cref{tab:table_1}, our method achieves only moderate improvements over Distill when the validation set size is large (10 samples per class). However, it demonstrates significant improvements when the validation set size is smaller, or the noise rate is higher (e.g., 11.5\% improvement for CIFAR-100 with 0.8 uniform noise and 4\% improvement for CIFAR-100 with 0.2 uniform noise, using 5 samples per class in the validation set). Similarily, our method demonstrates significant improvements on imbalanced and noisy benchmarks, as shown in ~\cref{tab:table_7}. This property highlights that our method performs better in scenarios with an extreme scarcity of clean data, especially when the dataset is both noisy and imbalanced. A possible explanation for this phenomenon is that when the number of clean training samples is limited, the informativeness of each sample becomes more critical for training compared to scenarios with abundant clean training data.

        \begin{table*}[ht]
        \centering
        \caption{Test accuracy (in \%) comparison between  our method (``INOLML'') and the Distill noise (``DN'') on symmetric noise using 1, 5 and 10 samples per class in the validation set on two backbone models: Resnet29 (``RN29'') and Wideresnet28-10 (``WRN''). The results of the Distill model with WideResnet28-10 are collected from \citep{Distill_noise}.
        Recall that the Distill needs a clean set, while INOLML works with a pseudo-clean set.}
        
        \label{tab:table_1}
        \scalebox{1.0}{
        \begin{tabular}{l c l  l  l  l  l  l}
            \toprule
            \multirow{3}{*}{\bfseries Method} & \multirow{3}{*}{\shortstack{\bfseries Val. Set\\ \bfseries size}} & \multicolumn{6}{c}{\bfseries Dataset}  \\
            \cmidrule(lr){3-8}
            & & \multicolumn{3}{c}{\bfseries CIFAR10} & \multicolumn{3}{c}{\bfseries CIFAR100}  \\
            \cmidrule(lr){3-5} \cmidrule(lr){6-8} 
            & & \multicolumn{1}{c}{\bfseries 0.2} & \multicolumn{1}{c}{\bfseries 0.4} & \multicolumn{1}{c}{\bfseries 0.8} & \multicolumn{1}{c}{\bfseries 0.2} & \multicolumn{1}{c}{\bfseries 0.4} & \multicolumn{1}{c}{\bfseries 0.8} \\
            \midrule
            DN-RN29  & \multirow{2}{*}{1} & 87.0 \(\pm\) 0.5 & 85.3 \(\pm\) 0.5 & \multicolumn{1}{c|}{FAIL} & 58.9 \(\pm\)  0.5 & 55.8 \(\pm\) 0.7 & \multicolumn{1}{c}{FAIL} \\
            INOLML-RN29  & & 90.3 \(\pm\) 0.2 & 89.1 \(\pm\) 0.5 & 79.1 \(\pm\) 0.3 & 65.9 \(\pm\) 0.2 & 61.5 \(\pm\) 0.4 & 55.1 \(\pm\) 0.6 \\
            \midrule
            DN-RN29  & \multirow{2}{*}{5} & 90.7 \(\pm\) 0.3 &89.0 \(\pm\) 0.3 & 83.5 \(\pm\) 0.2 & 62.6 \(\pm\)  0.4 & 58.8 \(\pm\) 0.5 & 48.5 \(\pm\) 0.5 \\
            INOLML-RN29  & & 90.9 \(\pm\) 0.2 & 90.9 \(\pm\) 0.1&87.4 \(\pm\)  0.2&66.6 \(\pm\)  0.1&65.7 \(\pm\)  0.1&59.0 \(\pm\) 0.5 \\
            \midrule
            DN-RN29  & \multirow{2}{*}{10} & 91.0 \(\pm\) 0.2 & 89.2 \(\pm\) 0.1&87.0 \(\pm\) 0.1&63.7 \(\pm\)  0.2&60.5 \(\pm\) 0.2&57.5  \(\pm\) 0.5 \\
            INOLML-RN29  & & 92.2 \(\pm\) 0.1 & 91.0 \(\pm\) 0.1 & 87.9 \(\pm\)  0.2 & 67.1 \(\pm\) 0.1 & 66.3 \(\pm\)  0.1 & 59.2 \(\pm\) 0.2 \\
            \midrule
            \midrule
            DN-WRN  & \multirow{2}{*}{1} & 95.4 \(\pm\) 0.6 & 94.5 \(\pm\) 1.0 & 87.9 \(\pm\) 5.1 & 77.4 \(\pm\) 0.4 & 75.5 \(\pm\) 1.1 & 62.1 \(\pm\) 1.2 \\
            INOLML-WRN  & & 96.0 \(\pm\) 0.2 &95.9 \(\pm\) 0.2&94.3 \(\pm\) 0.2&81.6 \(\pm\) 0.2&79.5 \(\pm\) 0.2&73.6 \(\pm\) 0.3 \\
            \midrule
            DN-WRN  & \multirow{2}{*}{5} & 96.4 \(\pm\) 0.0 & 95.5 \(\pm\) 0.6 & 91.8  \(\pm\)  3.0 & 80.4 \(\pm\) 0.5 & 79.6  \(\pm\)  0.3 & 73.6 \(\pm\) 1.5 \\
            INOLML-WRN & & 96.4 \(\pm\) 0.1 &96.2 \(\pm\) 0.1 & 94.6 \(\pm\) 0.2 & 82.2 \(\pm\) 0.2&81.5 \(\pm\) 0.2&74.5 \(\pm\)  0.2 \\
            \midrule
            DN-WRN & \multirow{2}{*}{10} & 96.2 \(\pm\) 0.2 &95.9 \(\pm\) 0.2&93.7  \(\pm\)  0.5&81.2 \(\pm\) 0.7&80.2  \(\pm\)  0.3 & \(\mathbf{75.5 \pm 0.2}\) \\
            INOLML-WRN & & \(\mathbf{96.9 \pm 0.1}\) & \(\mathbf{96.6 \pm 0.1}\) & \(\mathbf{95.0 \pm 0.2}\) & \(\mathbf{82.0 \pm 0.2}\) & \(\mathbf{81.3 \pm 0.2}\) & 74.7 \(\pm\) 0.1 \\
            \bottomrule
        \end{tabular}
        }
    \end{table*}

    % \subsection{Open-set noise}
    % \label{sec:additional_openset_noise_results}
        
    %     \begin{table}[t]
    %         \centering
    %         \caption{Comparison of our methods with the Distill in open-set noise setting~\citep{OpenSet_noise} using WideResnet28-10 model with samples pess for the clean subset. \cuong{Is there a typo for this caption?}}
    %         \label{tab:openset_noise}
    %         \begin{tabular}{l l l l}
    %             \toprule
    %             \multirow{2}{*}{\bfseries Method}& \multicolumn{3}{c}{\bfseries Test set}  \\ \cmidrule{2-4} 
    %             & \bfseries ImageNet \qquad \quad & \bfseries CIFAR100 \qquad \quad & \bfseries CIFAR100 + ImageNet  \\
    %             \midrule
    %             RoG~\citep{Distill_noise} &  83.4 & 87.1 & 84.4  \\
    %             L2R~\citep{Distill_noise} &  81.8 & 81.8 & 85.0  \\
    %             Distill~\citep{Distill_noise} &  94.0 & 92.3 & 93.0  \\
    %             \rowcolor{gray!30} \textbf{Ours} &  \textbf{94.5 \(\pm\) 0.1}  & \textbf{93.6 \(\pm\) 0.0} & \textbf{93.6 \(\pm\) 0.1} \\
    %             \bottomrule
    %         \end{tabular}
    %     \end{table}

\newpage
\section{Additional Results of Semantic Noise on CIFAR100 Dataset}
\label{sec:additional_semantic_noise_results}
        We provide additional results for semantic noise introduced by RoG~\citep{RoG}  comparing our proposed method and the Distill model~\citep{Distill_noise} in Table \ref{tab:table_semantic}. Both methods utilize the ResNet29 architecture, with validation set sizes ranging from 1, 5, to 10 samples per class.

        \begin{table}[H]
                \centering
                \caption{Test accuracy (\%) of INOLML on CIFAR100 with the semantic noise introduced by RoG~\citep{lee2019robust}, in comparison with Distill using a validation set $\mathcal{D}^{(v)}$ of sizes 1, 5 and 10 samples per class on Resnet29 model. The superscript \textsuperscript{T} indicates the need for clean validation sets.}
                \label{tab:table_semantic}
                
                \centering
                  \begin{tabular}{l c c}
                        \toprule
                        \bfseries Method & $|\mathcal{D}^{(v)}|$ & \bfseries Accuracy \\
                        \midrule
                        Distill\textsuperscript{T} & \multirow{2}{*}{$1\times C$} & 58.1 \(\pm\) 0.1   \\
                        \textbf{INOLML} & & 61.5 \(\pm\) 0.2  \\
                        \midrule
                        Distill\textsuperscript{T} & \multirow{2}{*}{$5\times C$} & 61.3 \(\pm\) 0.3  \\
                        \textbf{INOLML} & & 63.1 \(\pm\) 0.3  \\
                        \midrule
                        Distill\textsuperscript{T} & \multirow{2}{*}{$10\times C$} & 61.5 \(\pm\) 0.3 \\
                        \textbf{INOLML} & & \(\mathbf{63.5} \pm \mathbf{0.1}\) \\
                        \bottomrule
                    \end{tabular}
            \end{table}

\end{document}